%% file: main.tex
\documentclass[11pt,a4paper]{article}

\usepackage{polyu-vclab}

\usepackage{amsmath,amssymb}
\usepackage{booktabs}
\usepackage{multirow}
\usepackage{makecell}
\usepackage{adjustbox}
\usepackage{array}
\usepackage{colortbl}
\usepackage{arydshln}
\usepackage{tabularx}
\usepackage{comment}
\usepackage[authoryear,round]{natbib}
\setcitestyle{aysep={,},yysep={;}}
\renewcommand{\cite}{\citep}
\usepackage[capitalize]{cleveref}

\setEyebrow{PolyU VCLab\,\textbullet\,Preprint 2026}
\setReportTag{Visual Computing Lab\,\textperiodcentered\,The Hong Kong Polytechnic University}

\papertitle{Sparsity-Adaptive Mixture-of-Depth for \\ Efficient Vision-Language Models}

\paperauthors{%
  Bairui Li\affilmark{1,2}\quad
  Xinyu Wei\affilmark{1,2}\quad
  Xuyang Liu\affilmark{1,2}\quad
  Jinrui Zhang\affilmark{1,2}\quad
  Zhiyuan Qi\affilmark{1}\\[3pt]
  Zeqing Wang\affilmark{1,2}\quad
  Wentong Li\affilmark{1}\correspondmark$^,$\notemark\quad 
  Lei Zhang\affilmark{1,2}\correspondmark%
}

\paperaffil{%
  \affilmark{1}\,The Hong Kong Polytechnic University\quad
  \affilmark{2}\, OPPO Research Institute%
}

\papernotes{%
  \correspondmark\,Corresponding authors (\href{mailto:cslzhang@comp.polyu.edu.hk}{liwentong95@gmail.com, cslzhang@comp.polyu.edu.hk}). 
  \notemark\,Work done during visiting in PolyU.
}

\paperbadges{%
  \vclabbadgesolid{GitHub}{https://github.com/PolyU-VCLab/SA-MoD}\;%
  \vclabbadge{Model}{https://huggingface.co/VCLab-PolyU/SA-MoD}\;%
  \vclabbadge{Dataset}{https://huggingface.co/datasets/VCLab-PolyU/SA-MoD-41K}\;%
}

\begin{document}
\maketitleVCLab

\input{sec/0_abstract}

\input{sec/1_introduction}
\input{sec/2_related_works}
\input{sec/3_methodology}
\input{sec/4_experiments}
\input{sec/5_conclusion}

\bibliography{main}

\appendix
\input{sec/A_appendix}

\end{document}

%% file: sec/0_abstract.tex
\begin{vclabAbstract}
\noindent\textbf{Abstract.}\;
Vision-language models (VLMs) incur substantial computational costs during prefilling for long-context visual understanding. Many visual token reduction methods rely on fixed-budgets or predefined training schedules without sparsity-adaptivity, while content-adaptive approaches may entail train-inference discrepancies, complex training pipelines, or incompatibility with progressive sparsification.
We propose \textbf{Sparsity-Adaptive Mixture-of-Depth (SA-MoD)}, a single-stage post-training framework that progressively learns content-adaptive, layer-wise visual token routing through a dense-to-sparse transition, by integrating The Parameterized Thresholding Mechanism, RouterNorm, and Density-aware Regularization. We adopt On-Policy Distillation as the training objective and introduce Online Hidden-State Alignment (OHiSA), which aligns teacher and student hidden representations conditioned on the student's sparse trajectory, to provide reliable layer-wise supervision. Extensive experiments on Qwen3.5-4B across 12 image and 5 video understanding benchmarks demonstrate that SA-MoD retains \textbf{99.60\%} and \textbf{95.57\%} of the dense model's performance on image tasks with \textbf{25\%} and \textbf{5\%} vision-token retention, respectively, while preserving \textbf{97.78\%} and \textbf{96.77\%} of the dense model's performance on video tasks at \textbf{25\%} and \textbf{10\%} retention, respectively. For efficiency, SA-MoD achieves up to \textbf{3.95$\times$} end-to-end speedup. These results demonstrate the potential of SA-MoD for efficient VLM inference under long visual contexts. 
\end{vclabAbstract}

\vspace{-8pt}
\input{figs/amod_teaser}

%% file: figs/amod_teaser.tex
\begin{figure}[b]
    \centering
    \includegraphics[width=\linewidth]{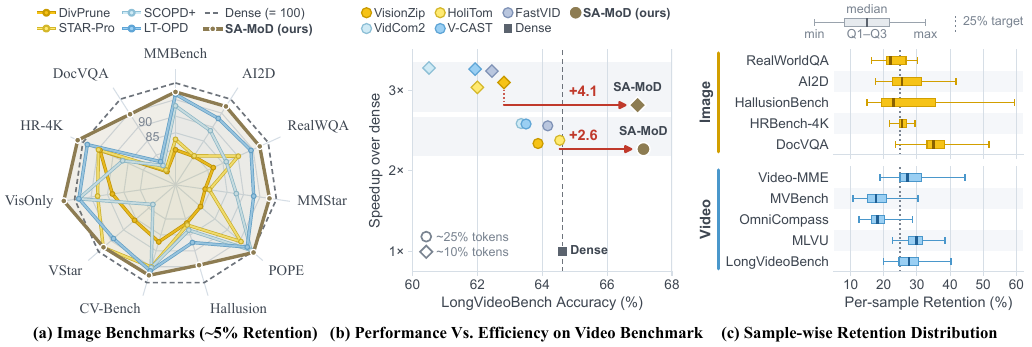}
    \caption{\textbf{SA-MoD achieves adaptive visual-token sparsification with strong performance--efficiency balance.}
    (a) Performance comparison at high sparsity: under 5\% retention, SA-MoD outperforms all prior methods and even exceeds the dense model on several benchmarks.
    (b) Performance-efficiency balance on video tasks: SA-MoD gains +2.6 and +4.1 points over its respective best competitor on LongVideoBench at 25\% and 10\% retention with comparable speedup.
    (c) Retention distribution on different benchmarks: under a 25\% retention target, sample-wise retention rates dynamically adapt to input complexity instead of adhering to a fixed ratio.}
    \label{fig:amod_teaser}
\end{figure}

%% file: sec/1_introduction.tex
\section{Introduction}
\label{sec:introduction}

Vision-language models (VLMs) have achieved remarkable progress in visual understanding and multimodal reasoning~\cite{qwen2026qwen35,comanici2025gemini,team2026kimi,xu2026deepseek}. However, processing high-resolution images and long videos introduces massive visual token sequences, resulting in substantial computational overhead during prefilling. As visual context grows, visual token processing becomes a major bottleneck for VLM inference~\cite{shao2025survey,yao2026towards,jin2025efficient}. 
Previous studies~\cite{fastv,sparsevlm,li2025tokenpacker,yang2025visionzip,ye2025atp} have revealed substantial redundancy among visual tokens, motivating the development of various token reduction methods. 

Existing visual token reduction approaches fall into two main paradigms: \textit{predefined sparsity} and \textit{adaptive sparsity}. The former includes training-free methods~\cite{divpruner,zoo-prune,yu2026visiontrim,guo2026star} that typically prune visual tokens using heuristic strategies under fixed Top-K budgets, and learning-based methods~\cite{wen2025efficient,jeddi2026scopd,li2026fewer-ltopd,zhang2025p,luo2025gamma,dynamic-llava}
that train the model under given pruning strategies or with additional learnable selection modules. However, these approaches impose fixed retention budgets across inputs of various visual complexity and information density.
Although some methods~\cite{wen2025efficient,li2026fewer-ltopd} introduce progressive curricula to enable smoother optimization, they still adopt manually prescribed schedules controlled with fixed budgets. Another line of
methods~\cite{ye2025atp,wang2025dymu,feng2026visionweave} has explored content-adaptive token allocation through learnable thresholds~\cite{ye2025atp} or probability-based routing~\cite{feng2026visionweave}.
Notably, these methods optimize token allocation towards predefined sparsity targets, lacking explicit progressive sparsity control throughout the training process.
Meanwhile, they mostly rely on surrogate gradients or differentiable relaxations (\textit{e.g.}, Gumbel-Softmax~\cite{jang2017categorical} or Straight-Through Estimators~\cite{ye2025atp,takezoe2026learnpruner}), leading to train-inference discrepancies and optimization instability, while some others require complex, multi-stage training pipelines in practice~\cite{feng2026visionweave}.

We conjecture that sparsity adaptiveness should be learned progressively through end-to-end optimization, rather than being imposed by predefined fixed budgets or manually prescribed training schedules. This process allows the model to gradually self-allocate computations according to the input visual complexity and informativeness, spontaneously yielding content-adaptive sparsity.
Towards this objective, we identify the Mixture-of-Depths (MoD) paradigm~\cite{raposo2024mixture,zhang2025p,dong2026xmodpracticalscalinglaws} as a promising architectural foundation, where tokens selectively bypass each Transformer layer. MoD can flexibly allocate sparsity across different depths of the architecture, allowing tokens to skip individual layers while remaining available for subsequent layer computation. Furthermore, its routing formulation naturally obviates surrogate gradients or differentiable relaxations, thereby ensuring strict consistency between training and inference.

Building on this insight, we present \textbf{Sparsity-Adaptive Mixture-of-Depth (SA-MoD)}, an end-to-end post-training framework that progressively learns layer-wise, content-adaptive visual token routing through a dense-to-sparse transition. 
Starting from a pretrained VLM, 
SA-MoD optimizes token routing under increasingly sparse computation to faithfully preserve the base model capabilities.
Specifically, a \textit{parameterized thresholding} mechanism is introduced to learn adaptive routing boundaries, while \textit{RouterNorm} is used to stabilize the optimization of layer-wise routing. To facilitate progressive sparsification, we further introduce \textit{density-aware regularization}, which provides constant gradients to gradually sparsify the model without manually prescribing step-wise or layer-wise sparsity schedules. 
To further preserve model performance under progressive sparsification, we adopt On-Policy Distillation (OPD) as the training objective~\cite{agarwal2024policy,li2026fewer-ltopd,jeddi2026scopd}, where a frozen dense teacher provides token-level supervision along the sparse model's rollout trajectories.
Based on OPD, we propose Online Hidden-State Alignment (OHiSA), which distills layer-wise teacher representations conditioned on the student's hidden-state inputs, providing fine-grained layer-wise supervision to preserve model performance under sparse routing.

We evaluate SA-MoD on Qwen3.5-4B~\cite{qwen2026qwen35} across 12 image and 5 video understanding benchmarks. On image tasks, SA-MoD retains \textbf{99.60\%} and \textbf{95.57\%} of the original model's performance with \textbf{25\%} and \textbf{5\%} visual tokens, respectively, demonstrating strong robustness under high token reduction ratios (see \cref{fig:amod_teaser}(a)). On video tasks, SA-MoD retains \textbf{97.78\%} and \textbf{96.77\%} of the original model's performance at \textbf{25\%} and \textbf{10\%} retention ratios, respectively. For efficiency, SA-MoD achieves up to \textbf{3.95$\times$} end-to-end speedup on long videos, approaching the speed of training-free methods while delivering substantially better performance (see \cref{fig:amod_teaser}(b)). Furthermore, \cref{fig:amod_teaser}(c) illustrates SA-MoD's sparsity adaptiveness across different benchmarks.

Our main contributions are summarized as follows:
\begin{itemize}[itemsep=2pt, topsep=1pt, parsep=1pt]
    \item We propose \textbf{SA-MoD}, an end-to-end, single-stage post-training framework for sparsity-adaptive visual token routing via a dense-to-sparse transition. It integrates parameterized thresholding to learn adaptive routing boundaries, RouterNorm to stabilize layer-wise optimization, and density-aware regularization to enable progressive sparsification without prescribed schedules.

    \item We introduce Online Hidden State Alignment (OHiSA) to complement On-Policy Distillation (OPD) with layer-wise hidden-state supervision along the student's sparse computational trajectory, effectively preserving model performance under sparse token routing.

    \item Extensive experiments on Qwen3.5-4B across 12 image and 5 video understanding benchmarks demonstrate the robust performance of SA-MoD under various retentions, and competitive efficiency of up to \textbf{3.95$\times$} end-to-end speedup on long video contexts, highlighting the potential for efficient long-context visual understanding.
\end{itemize}
\vspace{-3pt}

%% file: sec/2_related_works.tex
\section{Related Works}
\label{sec:related_works}

\textbf{Token Reduction for Efficient VLMs}.
Efforts to improve VLM efficiency span multiple orthogonal directions, such as model compression for on-device deployment~\cite{wang2024cloud}, parallel decoding of structured outputs~\cite{wang2026locateanything}, and input-side curation of visual evidence presented to multimodal agents~\cite{xu2026viewcraft}. In particular, vision token reduction has been widely studied to alleviate computational overhead by reducing redundant visual information and token computation~\cite{shao2025survey,yao2026towards,jin2025efficient}.
Existing approaches can be categorized into training-free and learning-based methods. Training-free methods primarily rely on heuristic strategies to prune redundant visual tokens without additional training~\cite{fastv,sparsevlm,divpruner,xing2024pyramiddrop,yang2025visionzip,zoo-prune,yu2026visiontrim,fang2026prune-iclr2026,guo2026star,jiang-etal-2026-v,chen2026one,yang2026reroutedontremoverecoverable}. FastV~\cite{fastv} leverages attention scores to prune visual tokens in early LLM layers, while SparseVLM~\cite{sparsevlm} sparsifies visual tokens via text-guided attention. DivPrune~\cite{divpruner} selects representative tokens by maximizing feature diversity, whereas STAR-Pro~\cite{guo2026star} adopts stage-wise token reduction  with progressive refinement. Similar strategies have been extended to video understanding by exploiting spatio-temporal token redundancy~\cite{tao2025dycoke,liu2025video,shao2026holitom,shen2026fastvid,lin2026v}.

Learning-based methods introduce trainable components or additional optimization to improve token compression and model adaptation
~\cite{hu2024epic,hu2024matryoshka,ye2025atp,zhang2025p,shang2025llava,li2025tokenpacker,zhang2025llava,takezoe2026learnpruner,takezoe2026learnprunerv2,Wang_2025_ICCV,zhang2025dyntokdynamiccompressionvisual}. 
EPIC~\cite{wen2025efficient} employs progressive token- and layer-wise schedule for consistency distillation to adapt visual token compression. TokenPacker~\cite{li2025tokenpacker} learns compact visual representation through coarse-to-fine projection. 
p-MoD~\cite{zhang2025p} introduces mixture-of-depths routing to selectively process visual tokens within LLM layers, which imposes layer-wise Top-$K$ budgets.
LearnPruner~\cite{takezoe2026learnpruner} combines learnable visual importance estimation at the projector with query-guided token selection within backbone. 
Beyond fixed-budget compression, several studies~\cite{ye2025atp,wang2025dymu,feng2026visionweave} have explored input-adaptive allocation.
ATP-LLaVA~\cite{ye2025atp} learns input-dependent pruning thresholds using budget-aware optimization, while VisionWeave~\cite{feng2026visionweave} enables content-adaptive routing between coarse- and fine-grained visual representations.
In contrast, 
our SA-MoD jointly learns content-adaptive, layer-wise routing and progressive sparsification without prescribed schedules.

\vspace{+1mm}
\noindent\textbf{On-Policy Distillation}.
Knowledge distillation is a common tool for building efficient models, \textit{e.g.}, transferring knowledge from large MLLMs to edge-device models~\cite{wang2024cloud} or distilling diffusion models for faster generation~\cite{guo2026phasewise}.
In vision language models, conventional off-policy distillation typically relies on fixed, labeled training sequences, introducing distribution mismatch during auto-regressive inference. VoCo-LLaMA~\cite{ye2025voco} employs attention distillation to learn compact visual representations. 
In contrast, on-policy distillation (OPD) mitigates distribution mismatch by supervising student-generated trajectories with teacher feedback~\cite{song2026survey,li2026rethinkingonpolicydistillationlarge,jang-etal-2026-stable}. 
GKD~\cite{agarwal2024policy} establishes this paradigm for language models, while OPSD~\cite{zhao2026self} extends it to self-distillation using privileged teacher contexts.
Beyond output-level supervision,  OPRD~\cite{yang2026oprd} explores representation-level distillation by aligning intermediate teacher and student states along the same rollouts. 
Recent studies~\cite{bousselham2026vold,yuan2026vision,li2026video,liu2026visualadvantageonpolicydistillationvisionlanguage,aniri2026opdvvisualonpolicyselfdistillation,wang2026vzeroselfimprovingmultimodalreasoning} have extended OPD to VLMs for fine-grained visual understanding and reasoning.
For visual token reduction, SCOPD~\cite{jeddi2026scopd} employs a full-context teacher to supervise student rollouts under sparse visual inputs, while 
LT-OPD~\cite{li2026fewer-ltopd} uses a progressive token-budget curriculum to improve VLM adaptation to extreme visual sparsity. 
Despite these advances, existing OPD methods for visual token reduction primarily focus on output-level supervision, while representation-level alignment under dynamic sparsity remains less explored.
In this work, we introduce Online Hidden-State Alignment (OHiSA),  
which conditions teacher targets on the student's online hidden-state inputs to align layer-wise transitions along its sparse computational trajectory in our SA-MoD.
\vspace{-6pt}

%% file: sec/3_methodology.tex
\section{Methodology}
\label{sec:methodology}

\subsection{Preliminaries}
\label{subsec:preliminaries}

Mixture-of-Depths (MoD)~\cite{raposo2024mixture} formulates selective token routing across Transformer layers in the LLM backbone. 
At each layer $l \in \{0, \dots, L-1\}$, let $I_l$ denote the set of input token indices, with hidden states $H_l = \{h_i\}_{i \in I_l}$. The MoD framework can be formalized into three key components:

\vspace{+2pt}
(1) \textsl{Router Module.} A lightweight router module $f_l$ maps each visual token state $h_i \in \mathbb{R}^D$ to routing signals. It employs a bias-free, zero-initialized linear projection to generate a scalar score $s_i \in \mathbb{R}$.

\vspace{+2pt}
(2) \textsl{Routing Criterion.} A binary decision $m_{i} \in \{0, 1\}$ determines whether token $i$ is processed by the current layer computation ($m_i = 1,~\mathrm{Keep}$) or bypasses it directly to layer $l+1$ ($m_i = 0,~\mathrm{Skip}$), remaining available for subsequent layers. MoD adopts a fixed-budget Top-$K$ criterion:
{
\setlength{\abovedisplayskip}{5pt}
\setlength{\belowdisplayskip}{5pt}
\begin{equation}
    m_i = \mathbb{I}\big(s_i - \mathrm{Sort}_{j \in I_l}(s_j)[K] \ge 0\big),~~i \in I_l,
    \label{eq:pmod_topk_equiv}
\end{equation}
}%
where $\mathbb{I}(\cdot)$ is the indicator function. Top-$K$ is equivalent to thresholding each $s_i$ against the $K$-th largest score.

\vspace{+2pt}
(3) \textsl{Branch Reweight}. A scalar gating factor $k_i$ modulates token representations and establishes differentiable gradient pathways for router optimization. Following MoD, $k_i = \sigma(s_i)$ scales the layer output on the $\mathrm{Keep}$ path, whereas the $\mathrm{Skip}$ path retains the identity mapping unchanged.

Building upon this basic formulation, we propose \textbf{SA-MoD}, which aims at visual token routing in general-purpose VLMs. To enable content-adaptive sparsity and progressive sparsification, we systematically redesign the router mechanism and optimization objectives.
\vspace{-4pt}

\subsection{SA-MoD Framework}
\label{subsec:samod_arch}
As illustrated in \cref{fig:pipeline}, SA-MoD follows a single-stage post-training framework that learns content-adaptive visual token routing through a dense-to-sparse transition. Specifically, we introduce \textit{Parameterized Thresholding} to enable content-adaptive sparsity, decoupling token importance scoring from sparsity control, \textit{RouterNorm} to mitigate hidden-state scale disparities to stabilize layer-wise router optimization, and \textit{Density-Aware Sparsity Regularization} to guide progressive sparsification without predefined schedules. During training, we integrate OPD \cite{agarwal2024policy} with \textit{Online Hidden-State Alignment}, preserving model capabilities under increasing sparsity with layer-wise supervision.

\input{figs/pipeline}

\vspace{+1mm}
\noindent\textbf{Parameterized Thresholding.}
We generalize the formulation of the \textit{Routing Criterion} in \cref{eq:pmod_topk_equiv} as:
{
\setlength{\abovedisplayskip}{5pt}
\setlength{\belowdisplayskip}{5pt}
\begin{equation}
    m_i = \mathbb{I}\big(S(h_i) + T(H_{l,v}) \ge 0\big),~~i \in I_{l,v},
    \label{eq:routing_criterion}
\end{equation}
}%
where $I_{l,v}$ and $H_{l,v}$ denote the indices and hidden states of visual entity $v$ (\textit{e.g.}, an image or video) at layer $l$, respectively. Here, $S(\cdot)$ represents the token-level scoring function, while $T(\cdot)$ denotes the entity-level routing threshold.

A simple approach to adaptiveness is \textit{Vanilla Thresholding}, which sets $T(H_{l,v}) \equiv \delta$. Unlike Top-$K$ routing, this formulation naturally enables content-adaptive sparsity with full-range token retention. However, the fixed threshold forces $S(\cdot)$ to simultaneously learn relative token importance (\textit{scoring}) and suppress absolute magnitudes to control sparsity (\textit{sifting}),  inducing gradient conflicts and optimization instability, particularly under increasingly low retention.

To decouple \textit{scoring} from \textit{sifting}, we propose \textit{Parameterized Thresholding}, which leverages a linear router with bias $f_l: \mathbb{R}^D \to \mathbb{R}^2$ to project each token $h_i$ into a score component $s_i$ and a threshold component $t_i$. The threshold function $T(\cdot)$ and the scoring function $S(\cdot)$ are defined as:
{
\setlength{\abovedisplayskip}{6pt}
\setlength{\belowdisplayskip}{6pt}
\begin{align}
    &[s_i,\; t_i]^\top = f_l(h_i) = W_l h_i + b_l,~~~~~~~i \in I_{l,v}, \label{eq:router_proj} \\
    &T(H_{l,v}) = \mathrm{mean}_{j \in I_{l,v}}(t_j) + \delta,~~S(h_i) = s_i,
    \label{eq:param_threshold}
\end{align}
}%
with a constant margin $\delta > 0$. Notably, mean-pooling the threshold component $t_j$ over $I_{l,v}$ establishes a shared entity-level boundary $T(H_{l,v})$ across all visual tokens in the input. This decoupled design preserves the relative token salience encoded in $s_i$ without distorting intra-image importance rankings, while allowing the router to adaptively regulate overall retention rates.

For branch reweighting, we adopt the \textit{TanhNorm} and \textit{STRing} mechanisms from $p$-MoD~\cite{zhang2025p}. Letting $\hat{h}^{(l)}_i = h_i^{(l)} + \mathrm{Attn}_l(h_i^{(l)})$ denote the post-attention representation, the reweighting factor $k_i$ and the hidden-state update rule are defined as:
{
\setlength{\abovedisplayskip}{5pt}
\setlength{\belowdisplayskip}{2pt}
\begin{align}
    &k_i = 1 + w\tanh\!\left(S(h_i) + T(H_{l,v}) - \delta\right),~~i \in I_{l,v}, 
    \label{eq:samod_reweight} \\
    &h_i^{(l+1)} = \begin{cases}
    \hat h^{(l)}_i + k_i \cdot \mathrm{MLP}_l(\hat h^{(l)}_i), & ~~~~~\text{if } m_i = 1, \\
    k_i \cdot h_i^{(l)}, & ~~~~~\text{if } m_i = 0.
    \end{cases}
    \label{eq:mod_branch}
\end{align}
}%

\vspace{+2mm}
\noindent\textbf{RouterNorm: Pre-Router Mean RMSNorm.}
Hidden-state norms can vary substantially across Transformer layers and VLM backbones\footnotemark[1], leading to imbalanced router optimization.
For a linear router $y = W h + b$, the gradient with respect to its weight matrix is $\frac{\partial \mathcal{L}}{\partial W} = \frac{\partial \mathcal{L}}{\partial y}h^\top$. Therefore, the weight update $\Delta W = -\eta\frac{\partial \mathcal{L}}{\partial W}$ induces the score to shift \textit{quadratically} with the input norm: $\Delta y = \Delta W h = -\eta\frac{\partial \mathcal{L}}{\partial y}\|h\|^2 \propto \|h\|^2$.
Under a uniform margin $\delta$, this would cause premature sparsification in high-norm layers and sluggish updates followed by abrupt retention collapse in low-norm layers. 

To address this, we introduce \textit{RouterNorm}, a parameter-free \textit{Pre-Router Mean RMSNorm} that normalizes hidden states per layer $l$ and entity $v$ before router projection:
{
\setlength{\abovedisplayskip}{5pt}
\setlength{\belowdisplayskip}{5pt}
\begin{equation}
    h_i \gets \frac{h_i}{\mathrm{mean}_{j \in I_{l,v}}\big[\mathrm{sg}\big(\sqrt{\tfrac{1}{D}\|h_j\|^2}\big)\big]},~~i\in I_{l,v},
    \label{eq:routernorm}
\end{equation}
}%
where $D$ is the hidden size and $\mathrm{sg}(\cdot)$ denotes the stop-gradient operator. 
\textit{RouterNorm} ensures cross-layer scale consistency and training stability while strictly preserving intra-entity token relative importance.

\footnotetext[1]{Different attention architectures, such as full attention and linear attention, can exhibit distinct hidden-state scale distributions.}

\vspace{+1mm}
\noindent\textbf{Density-Aware Sparsity Regularization.}
\label{subsec:density_reg}
To control sparsity without a prescribed schedule, we apply a regularization loss $\mathcal{L}_{\text{Reg}}$ that directly penalizes the threshold $T(H_{l,v})$, which suppresses the threshold so that low-scoring tokens gradually fall below the routing boundary (\cref{eq:routing_criterion}), while leaving relative scoring $S(h_i)$ undisturbed. To start training, the router $(W_l, b_l)$ is zero-initialized so that $m_i \equiv k_i \equiv 1$ in \cref{eq:routing_criterion,eq:param_threshold,eq:samod_reweight}, which ensures a dense initialization equivalent to the base model.
In early steps, the positive margin $\delta$ creates a buffer zone $[0, \delta]$, keeping full retention to establish reliable token importance. As $T(H_{l,v})$ decays, less informative tokens are progressively skipped until reaching a dynamic equilibrium between sparsity and performance.
However, static regularization, which reduces the threshold at a constant strength, risks abrupt retention collapse if tokens cluster near the decision boundary, or updates sluggishly when most tokens remain safely above the threshold.

To address this issue, we propose \textit{Density-Aware Regularization}, which adaptively adjusts regularization strength according to the proximity of kept tokens to the decision boundary. We first define the density-aware coefficient as:
{
\setlength{\abovedisplayskip}{5pt}
\setlength{\belowdisplayskip}{5pt}
\begin{equation}
    \lambda_{l,v} = \mathrm{mean}_{i\in I_{l,v}}\!\left[\mathrm{ReLU}\!\left(\frac{\tanh\!\big(\frac{1}{\delta}\mathrm{sg}(\Delta_{l,v,i})\big)}{\tanh(1)}\right)\right], \label{eq:density_w}
\end{equation}
}%
where $\Delta_{l,v,i} = S(h_{l,v,i}) + T(H_{l,v})$ denotes the decision margin of token $i \in I_{l,v}$ as defined in \cref{eq:routing_criterion}.

Specifically, $\lambda_{l,v}$ in \cref{eq:density_w} measures the averaged distance of kept tokens from the decision boundary, $i.e.$, \textit{near-boundary token density}, acting as an automatic step-size controller: 
(1) When tokens cluster near the boundary ($\Delta_{l,v,i} \to 0^+$), $\lambda_{l,v}$ shrinks to throttle threshold decay and prevent mass decision flips; 
(2) With most tokens safely above the boundary ($\Delta_{l,v,i} \gg 0$), $\lambda_{l,v} \uparrow$ safely accelerates sparsification; and 
(3) Skipped tokens ($\Delta_{l,v,i} < 0$) are truncated by $\mathrm{ReLU}$ without unnecessary gradients.

The density-aware regularization loss is defined as:
{
\setlength{\abovedisplayskip}{5pt}
\setlength{\belowdisplayskip}{5pt}
\begin{equation}
    \mathcal{L}_{\text{DReg}} = \mathbb{I}(R>\hat R)\cdot \mathrm{mean}_{l,v}\!\left[ \lambda_{l,v}\cdot T(H_{l,v}) \right]
    \label{eq:l_reg}, 
\end{equation}
}%
where the threshold is weighted by a density-aware coefficient and averaged across layers and visual entities. Once the global retention rate $R$ (defined as the average visual token retention ratio across all layers for the current sample) reaches the target retention $\hat{R}$, the regularization deactivates to prevent over-sparsification and steer the retention convergence.

\vspace{-2pt}
\subsection{Online Hidden-State Alignment}
\label{subsec:ohisa}
Task-loss gradients tend to diminish as they back-propagate to early layers, causing insufficient router optimization and calling for auxiliary layer-wise supervision.
However, conventional layer-wise distillation methods~\cite{yang2026oprd} directly align the corresponding student-teacher hidden states: $H_l^A \rightarrow H_l^T$, suffering from cumulative state divergence, particularly under layer-wise sparsification.

To address this, we propose \textit{Online Hidden-State Alignment} (OHiSA). At each layer $l$, $H_l^A$ and $H_l^T$ denote the student and teacher hidden states, with $F_l^A$ and $F_l^T$ being the student and teacher layer blocks, respectively. Instead of referencing a drifted teacher state $H^T_{l+1}=F_l^T(H_l^T)$, we feed the student's current state $H_l^A$ directly into the corresponding teacher block $F_l^T$ to construct the \textit{Online Teacher Target}: $H_{l+1}^{T,\mathrm{on}} = F_l^T\!\left(\mathrm{sg}(H_l^A)\right)$, with stop-gradient applied to $H_l^A$.

We minimize the layer-averaged mean squared error (MSE) between the student's output $H_{l+1}^A = F_l^A(H_l^A)$ and $H_{l+1}^{T,\mathrm{on}}$ over the response tokens $\mathcal{R}$ at every layer:
{
\setlength{\abovedisplayskip}{1pt}
\setlength{\belowdisplayskip}{3pt}
\begin{equation}
    \mathcal{L}_{\text{OHiSA}} = \frac{1}{L \cdot |\mathcal{R}|}\sum_{l=0}^{L-1}\sum_{j \in \mathcal{R}} \left\| H_{l+1,j}^A - H_{l+1,j}^{T,\mathrm{on}} \right\|_2^2.
    \label{eq:loss_on}
\end{equation}
}%
By conditioning the teacher and student transitions on the same input state $H_l^A$, OHiSA provides layer-wise reliable supervision for router optimization.

\vspace{-2pt}
\subsection{Training Objectives}
\label{subsec:training_objectives}

\noindent\textbf{On-Policy Distillation.}
We optimize SA-MoD via single-stage OPD~\cite{agarwal2024policy}, with the sparse SA-MoD model as the \textit{Student} $\pi_\theta$ and the frozen base model as the \textit{Teacher} $\pi_{T}$. Given an input $x$, the student rolls out a sequence $y \sim \pi_\theta(\cdot \mid x)$, and we minimize the average token-level Jensen-Shannon Divergence (JSD)~\cite{fuglede2004jensen} between the teacher and student output logits on the response tokens $\mathcal{R}$:
{
\setlength{\abovedisplayskip}{4pt}
\setlength{\belowdisplayskip}{3pt}
\begin{equation}
    \mathcal{L}_{\text{OPD}} = \frac{1}{|\mathcal{R}|}\sum_{j \in \mathcal{R}} \mathrm{JSD}\big(\pi_{T}(\cdot | x, y_{<j}), \pi_\theta(\cdot | x, y_{<j})\big).
    \label{eq:loss_opd}
\end{equation}
}%
An effective OPD optimization requires high top-$K$ logit overlap between Teacher and Student under the same prefix~\cite{li2026rethinkingonpolicydistillationlarge}. By initializing SA-MoD with the dense state ($\pi_\theta \equiv \pi_{T}$) and progressively sparsifying under $\mathcal{L}_{\text{DReg}}$, it keeps strong distributional overlap throughout training.

\vspace{+1mm}
\noindent\textbf{Overall Objective.} Combining the objectives in \cref{eq:l_reg}, \cref{eq:loss_on}, and \cref{eq:loss_opd}, with $\lambda_{\text{DReg}}$ and $\lambda_{\text{OHiSA}}$ balancing the density-aware sparsity regularization and online hidden-state alignment respectively, the overall training loss is formulated as:
$\mathcal{L}_{\text{total}} = \mathcal{L}_{\text{OPD}} + \lambda_{\text{DReg}}\mathcal{L}_{\text{DReg}} + \lambda_{\text{OHiSA}}\mathcal{L}_{\text{OHiSA}}$.

%% file: figs/pipeline.tex
\begin{figure}[t]
    \centering
    \includegraphics[width=\linewidth]{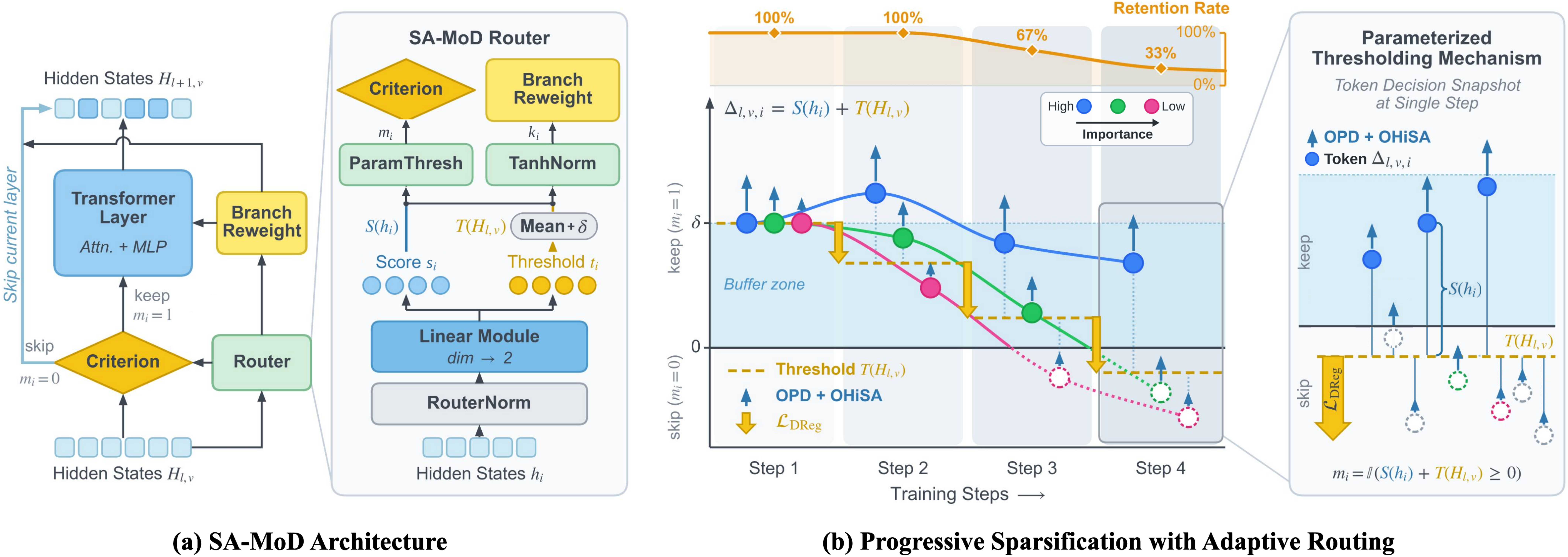}
    \caption{\textbf{Overview of SA-MoD Framework.}
    (a) SA-MoD performs visual token routing across Transformer layers. Integrated with parameterized thresholding and RouterNorm, the Router calculates token-wise scores $S$ and entity-level threshold $T$, thereby determining the criterion and branch-reweight.
    (b) During training, the density-aware regularization drives the dense-to-sparse transition, while OPD and OHiSA preserve model capabilities. The inset shows how opposing forces balance under parameterized thresholding: OPD and OHiSA retain important tokens ($m_i=1$), while $\mathcal{L}_{\text{DReg}}$ lowers $T(H_{l,v})$ so that unimportant ones are gradually skipped ($m_i=0$). The resulting routing decisions progressively adapt to token importance and reach the target retention rate, without fixed budgets or predefined schedules.}
    \label{fig:pipeline}
    \vspace{-6pt}
\end{figure}

%% file: sec/4_experiments.tex
\section{Experiments}
\label{sec:experiments}

\subsection{Experimental Setup}
\label{subsec:exp_settings}

\noindent\textbf{Compared Methods.}
We adopt Qwen3.5-4B~\cite{qwen2026qwen35} as the backbone. 
On both image and video domains, we compare SA-MoD with two recent open-source \textit{learning-based} token reduction methods: SCOPD+~\cite{jeddi2026scopd} and LT-OPD~\cite{li2026fewer-ltopd}, and multiple \textit{training-free} baselines. 
We benchmark against FastV~\cite{fastv}, SparseVLM~\cite{sparsevlm}, DivPrune~\cite{divpruner}, ZOO-Prune~\cite{zoo-prune}, EntropyPrune~\cite{wang-etal-2026-entropyprune} and STAR-Pro~\cite{guo2026star}. 
In video spatio-temporal token reduction methods, our comparison includes DyCoke~\cite{tao2025dycoke}, VisionZip~\cite{yang2025visionzip}, HoliTom~\cite{shao2026holitom}, FastVID~\cite{shen2026fastvid}, VidCom$^2$~\cite{liu2025video} and V-CAST~\cite{lin2026v}.

\vspace{+1mm}
\noindent\textbf{Training Datasets.}
We train SA-MoD on two curated datasets: (1) \textit{SA-MoD-Image-25K} comprises 25K instances sampled from Honey-Data-15M~\cite{zhang2026bee}, covering general, OCR \& document, STEM, and grounding tasks.
(2) \textit{SA-MoD-Video-16K} comprises 16K video QA instances, sampled from Video-R1-260K following Video-R1~\cite{feng2025videor1reinforcingvideoreasoning} and CaCoVID~\cite{ma2026contributionawaretokencompressionefficient}. More dataset statistics and preprocessing settings are detailed in the Appendix \cref{appendix:exp_details}.

\vspace{+1mm}
\noindent\textbf{Benchmarks.}
We conduct comprehensive evaluation across a broad suite of 12 image-based benchmarks and 5 video benchmarks. For \textit{image} tasks, we cover four categories: (1)~fine-grained vision-centric: V$^*$Bench~\cite{wu2024v}, HRBench4K~\cite{HRBench4k}, CV-Bench~\cite{tong2024cambrian} and RealWorldQA~\cite{ai2024grok}; (2)~general VQA \& reasoning: MMBench$_{\text{V1.1}}$~\cite{liu2024mmbench}, MMStar~\cite{chen2024we}, and VisOnly~\cite{kamoi2024visonlyqa}; (3)~hallucination \& faithfulness: POPE~\cite{li2023evaluating} and HallusionBench~\cite{guan2024hallusionbench}; and (4) text-rich \& OCR: AI2D~\cite{kembhavi2016diagram}, OCRBench~\cite{liu2024ocrbench} and DocVQA~\cite{mathew2021docvqa}. For \textit{video} domain, we use Video-MME~\cite{fu2025video}, MVBench~\cite{li2024mvbench} and OmniCompass~\cite{liu2026omnicompass} for video understanding; MLVU~\cite{zhou2025mlvu} and LongVideoBench~\cite{wu2024longvideobench} for long-form video tasks. 
We report the task metrics with average performance retention (\textit{Avg.\ Perf.}) and average token retention (\textit{Avg.\ Retention}).\footnote[2]{%
\textit{Avg.\ Perf.:} $\mathrm{mean}_{i}(\text{Acc}_i / \text{Acc}_i^{\text{base}}) \times 100\%$ across benchmarks; \textit{Avg.\ Retention:} $\mathrm{mean}_{l}(N_l / N_{\text{total}}) \times 100\%$ across backbone layers ($N_l$: kept tokens at layer $l$; $N_{\text{total}}$: total visual tokens), and then averaged across benchmarks.}

\vspace{+1mm}
\noindent\textbf{Implementation Details.}
SA-MoD is trained on \textit{SA-MoD-Image-25K} and \textit{SA-MoD-Video-16K} for image and video adaptation, respectively.
Under OPD settings, we train SA-MoD in a single stage, with a constant learning rate of $2\times 10^{-6}$ and a warmup steps of 10. The global batch size is 32. We adopt 4 rollouts per sample with a temperature of 1.0 and a max response length of 256. We use the frozen base model as the teacher, with training steps spanning from 500 to 4,000 according to the target retention rate.
For a fair comparison, the learning-based methods (SCOPD+~\cite{jeddi2026scopd} and LT-OPD~\cite{li2026fewer-ltopd}) are trained on our image and video datasets, which outperform their original datasets (shown in Appendix \cref{tab:supp_datasets_compare}).

\vspace{-4pt}
\subsection{Main Results}
\label{subsec:main_results}

\noindent\textbf{Quantitative Comparison.}
We evaluate SA-MoD across both image and video understanding  benchmarks.
On image benchmarks (\cref{tab:main_results}), SA-MoD consistently achieves the highest average performance retention across all token budget settings. In particular, with only $5.18\%$ of the visual tokens retained, SA-MoD preserves an impressive $95.57\%$ of the original dense model's performance, substantially outperforming LT-OPD~\cite{li2026fewer-ltopd} ($89.84\%$), SCOPD+~\cite{jeddi2026scopd} ($81.13\%$) and STAR-Pro~\cite{guo2026star} ($78.49\%$).
Similar observations can be made on video benchmarks (\cref{tab:video_results}), where SA-MoD reaches $97.78\%$ and $96.77\%$ average performance retention at $25.89\%$ and $10.76\%$ visual token retention, respectively, significantly outperforming all evaluated baselines. 
The qualitative analysis, training dynamics and deployment details of SA-MoD are provided in the Appendix \cref{appendix:exp_details} and \cref{appendix:qualitative}.

\input{tabs/main_results}

\vspace{+1mm}
\noindent\textbf{Robustness across Retention Rates.}
\cref{fig:robustness} and \cref{tab:main_results} show that, whereas competing baselines degrade rapidly under tighter budgets, SA-MoD's performance declines smoothly as the token retention rate decreases, exhibiting no sharp accuracy drop even at the lowest 5\% retention level.

\vspace{+1mm}
\noindent\textbf{Cross-Architecture and Scale Adaptation.}
We extend SA-MoD to Qwen3.5-9B and to other VLM architectures beyond Qwen3.5~\cite{qwen2026qwen35}, including Qwen3-VL~\cite{bai2025qwen3} and LLaVA-OneVision-1.5~\cite{an2025llava}. As shown in \cref{tab:scales_arch}, SA-MoD preserves most of the performance on all these backbones, indicating its effectiveness across diverse model scales and families.

\input{figs/amod_sparsity}

\subsection{Efficiency Analysis}
\label{subsec:efficiency}

\noindent\textbf{Efficiency under Long Video Context.}
We evaluate SA-MoD's efficiency on LongVideoBench, measured with a per-GPU batch size of 1 in BF16. As shown in \cref{tab:main_efficiency}, SA-MoD consistently reduces inference costs across different video context lengths. With only $11.90\%$--$13.06\%$ average token retention, it achieves up to $3.95\times$ latency speedup and $3.97\times$ TTFT speedup, while reducing TFLOPs and KV-cache usage by up to $81.7\%$ and $85.5\%$, respectively. Notably, SA-MoD outperforms the dense baseline in all evaluated settings.

\input{tabs/video_results}
\input{tabs/main_efficiency}

\vspace{+1mm}
\noindent\textbf{Pareto Frontier of Efficiency and Performance.}
\cref{fig:amod_teaser}(b) compares LongVideoBench accuracy against end-to-end latency speedup over the dense baseline across token reduction methods, evaluated at 2.0 FPS with a maximum of 256 frames. While existing methods fall below the dense baseline in accuracy at both ${\sim}25\%$ and ${\sim}10\%$ token retention, SA-MoD surpasses the dense baseline under both retention levels, outperforming the competitive baseline by $+2.6\%$ and $+4.1\%$ at comparable speedups, respectively.

\subsection{Ablation Studies}
\label{subsec:ablation}
We conduct ablation studies to assess the contributions of \textit{sparsity criterion}, \textit{router design} and \textit{training objectives} to SA-MoD's performance at about $25\%$ retention. Summary metrics below are calculated over 8 image benchmarks; per-benchmark scores are reported in the Appendix \cref{appendix:full_bench}.

\vspace{+1mm}
\noindent\textbf{Progressiveness and Content-Adaptivity.}
To isolate the contributions of progressive sparsification and content-adaptive allocation, we conduct controlled comparisons within the unified MoD backbone using identical training recipes and comparable token budgets (\cref{tab:ablation_prs_ca}). Under this setup, SA-MoD achieves $100.26\%$ average performance retention, outperforming fixed Top-$K$ ($90.74\%$), fixed Top-$P$ ($89.61\%$), and curriculum Top-$K$ ($97.54\%$), which confirms that combining progressive sparsity decay with content-adaptive routing outperforms other settings.

\vspace{+1mm}
\noindent\textbf{Router Design.}
At similar token retention rates, using a fixed threshold causes the largest drop in mean performance retention (100.26\% to 91.15\%), highlighting the effectiveness of decoupling token scoring from sparsity control. Removing the density-aware coefficient or RouterNorm lowers performance to 98.32\% and 98.02\%, respectively (\cref{tab:ablation_amod_router}).

\input{tabs/ablation_panel}

\vspace{+1mm}
\noindent\textbf{Training Strategy.}
As shown in \cref{tab:ablation_training_strategy}, with the routing mechanism and OHiSA held fixed, OPD achieves the highest performance retention (100.26\%), followed by GRPO (98.28\%), off-policy distillation (OffPD) (96.02\%), and supervised fine-tuning (SFT) (94.68\%). These results highlight the advantage of on-policy optimization over off-policy supervision under sparse routing.

\vspace{+1mm}
\noindent\textbf{Ablation on OHiSA.}
As shown in \cref{tab:ablation_ohisa}, adding OHiSA raises average performance from $99.67\%$ (pure OPD without hidden-state supervision) to $100.26\%$, with consistent gains on text-heavy and general benchmarks such as RealWorldQA and DocVQA. In contrast, replacing OHiSA with naive layer-wise distillation~\cite{yang2026oprd} degrades performance to $96.99\%$, indicating that direct token-wise feature matching is overly restrictive under dynamic routing.

%% file: tabs/main_results.tex
\newcommand{\mainresultsfree}{\textcolor[HTML]{F6C317}{\small$\blacktriangle$\ }}
\newcommand{\mainresultstrain}{\textcolor[HTML]{2998EC}{\small$\bullet$\ }}
\begin{table}[htbp]
\centering
\caption{\textbf{Comparison of different visual token reduction methods on Qwen3.5-4B across image benchmarks and token retention rates.} 
\mainresultsfree and \mainresultstrain indicate training-free methods and learning-based methods, respectively.}
\label{tab:main_results}
\setlength{\tabcolsep}{3.0pt}
\begin{adjustbox}{max width=\textwidth}
\begin{tabular}{lccccccccccccc}
\toprule
\multirow{2}{*}{\textbf{Method}} & \multicolumn{3}{c}{\textbf{General VQA}} & \multicolumn{4}{c}{\textbf{Vision-Centric}} & \multicolumn{2}{c}{\textbf{Hallucination}} & \multicolumn{3}{c}{\textbf{Document}} & \multirow{2}{*}{\textbf{Avg. Perf. (\%)}} \\
\cmidrule(lr){2-4} \cmidrule(lr){5-8} \cmidrule(lr){9-10} \cmidrule(lr){11-13}
 & \textbf{MMB$_{\text{V1.1}}$} & \textbf{MMStar} & \textbf{VisOnly} & \textbf{V$^*$Bench} & \textbf{CV-Bench} & \textbf{RealWQA} & \textbf{HR-4K} & \textbf{POPE} & \textbf{Hallusion} & \textbf{AI2D$_{\text{test}}$} & \textbf{OCRB} & \textbf{DocVQA} & \\
\midrule
\multicolumn{14}{c}{\cellcolor[HTML]{EFF8FC}\textit{Dense Base Model (100\% Retention)}} \\
Qwen3.5-4B & 82.73 & 62.06 & 47.91 & 80.10 & 86.87 & 75.42 & 74.63 & 87.98 & 71.50 & 83.45 & 85.60 & 95.24 & 100\% \\
\midrule

\multicolumn{14}{c}{\cellcolor[HTML]{EFF8FC}\textit{Avg. Retention $\sim$51.10\% Visual Tokens}} \\
FastV \scriptsize{(ECCV'24)} \mainresultsfree & 79.02 & 56.53 & 46.08 & 80.62 & 85.86 & 73.59 & 76.13 & 86.26 & 67.09 & 75.97 & 50.10 & 86.36 & 92.83\% \scriptsize{\textcolor{gray}{(-7.17\%)}} \\
SparseVLM \scriptsize{(ICML'25)} \mainresultsfree & 82.05 & 61.86 & 47.48 & 79.58 & 86.70 & 76.33 & 76.13 & 87.81 & 70.07 & 82.38 & 62.20 & 93.11 & 97.27\% \scriptsize{\textcolor{gray}{(-2.73\%)}} \\
DivPrune \scriptsize{(CVPR'25)} \mainresultsfree & 81.40 & 60.53 & 47.73 & 79.58 & 86.17 & 73.07 & 76.63 & 87.78 & 67.50 & 81.54 & 74.40 & 90.48 & 97.29\% \scriptsize{\textcolor{gray}{(-2.71\%)}} \\
ZOO-Prune \scriptsize{(CVPR'26)} \mainresultsfree & 81.50 & 61.60 & 47.83 & 79.58 & 86.05 & 72.81 & 75.25 & 88.05 & 68.55 & 82.25 & 75.80 & 91.08 & 97.67\% \scriptsize{\textcolor{gray}{(-2.33\%)}} \\
EntropyPrune \scriptsize{(EMNLP'26)} \mainresultsfree & 80.19 & 61.06 & 47.58 & 79.05 & 85.97 & 73.72 & 73.75 & 87.17 & 68.98 & 81.63 & 72.00 & 90.84 & 96.81\% \scriptsize{\textcolor{gray}{(-3.19\%)}} \\
STAR-Pro \scriptsize{(Arxiv'2609)} \mainresultsfree & 82.04 & 62.00 & 48.70 & 80.62 & 86.71 & 76.86 & 75.25 & 88.53 & 70.36 & 82.30 & 66.90 & 92.92 & 98.11\% \scriptsize{\textcolor{gray}{(-1.89\%)}} \\
\noalign{\vskip\aboverulesep}
\hdashline
\noalign{\vskip\belowrulesep}
SCOPD+ \scriptsize{(Arxiv'2609)} \mainresultstrain & 82.23 & 64.03 & 49.44 & 78.53 & 87.08 & 74.24 & 72.38 & 88.90 & 71.02 & 82.17 & 77.90 & 94.71 & 99.05\% \scriptsize{\textcolor{gray}{(-0.95\%)}} \\
LT-OPD \scriptsize{(Arxiv'2609)} \mainresultstrain & 81.26 & 61.13 & 49.83 & 80.10 & 85.63 & 75.29 & 75.25 & 85.22 & 70.03 & 82.41 & 74.10 & 85.61 & 97.50\% \scriptsize{\textcolor{gray}{(-2.50\%)}} \\
\rowcolor[HTML]{FFF9D9} \textbf{SA-MoD (Ours)} \mainresultstrain & \textbf{81.95} & \textbf{63.36} & \textbf{52.21} & \textbf{81.55} & \textbf{87.37} & \textbf{77.18} & \textbf{76.13} & \textbf{88.58} & \textbf{70.05} & \textbf{82.19} & \textbf{78.90} & \textbf{94.84} & \textbf{100.48\%} \scriptsize{\textbf{\textcolor{blue!90!black}{(+0.48\%)}}} \\
\midrule

\multicolumn{14}{c}{\cellcolor[HTML]{EFF8FC}\textit{Avg. Retention $\sim$24.87\% Visual Tokens}} \\
FastV \scriptsize{(ECCV'24)} \mainresultsfree & 64.01 & 44.13 & 43.78 & 75.39 & 80.41 & 65.35 & 75.97 & 75.46 & 60.35 & 65.67 & 14.50 & 46.26 & 77.45\% \scriptsize{\textcolor{gray}{(-22.55\%)}} \\
SparseVLM \scriptsize{(ICML'25)} \mainresultsfree & 78.48 & 56.53 & 44.59 & 81.15 & 85.33 & 74.50 & 76.25 & 86.82 & 66.98 & 75.64 & 31.40 & 78.85 & 90.17\% \scriptsize{\textcolor{gray}{(-9.83\%)}} \\
DivPrune \scriptsize{(CVPR'25)} \mainresultsfree & 79.41 & 56.34 & 45.53 & 80.05 & 85.11 & 69.41 & 76.13 & 86.33 & 64.56 & 78.20 & 56.20 & 75.85 & 91.76\% \scriptsize{\textcolor{gray}{(-8.24\%)}} \\
ZOO-Prune \scriptsize{(CVPR'26)} \mainresultsfree & 79.17 & 56.93 & 46.69 & 76.96 & 85.20 & 71.37 & 74.88 & 87.72 & 64.35 & 76.61 & 51.30 & 73.51 & 91.05\% \scriptsize{\textcolor{gray}{(-8.95\%)}} \\
EntropyPrune \scriptsize{(EMNLP'26)} \mainresultsfree & 78.69 & 56.60 & 47.48 & 78.68 & 84.91 & 70.75 & 72.24 & 86.43 & 64.88 & 76.47 & 41.90 & 70.20 & 89.61\% \scriptsize{\textcolor{gray}{(-10.39\%)}} \\
STAR-Pro \scriptsize{(Arxiv'2609)} \mainresultsfree & 80.80 & 59.73 & 47.82 & 80.10 & 86.30 & 74.90 & 74.75 & 88.36 & 68.66 & 80.44 & 53.00 & 86.44 & 94.84\% \scriptsize{\textcolor{gray}{(-5.16\%)}} \\
\noalign{\vskip\aboverulesep}
\hdashline
\noalign{\vskip\belowrulesep}
SCOPD+ \scriptsize{(Arxiv'2609)} \mainresultstrain & 81.96 & 63.80 & 48.02 & 75.39 & 86.43 & 73.46 & 74.00 & 88.48 & 69.91 & 81.48 & 60.60 & 89.29 & 96.06\% \scriptsize{\textcolor{gray}{(-3.94\%)}} \\
LT-OPD \scriptsize{(Arxiv'2609)} \mainresultstrain & 81.19 & 58.93 & 49.07 & 78.53 & 85.60 & 74.64 & 74.13 & 85.47 & 68.24 & 81.44 & 68.60 & 78.66 & 95.28\% \scriptsize{\textcolor{gray}{(-4.72\%)}} \\
\rowcolor[HTML]{FFF9D9} \textbf{SA-MoD (Ours)} \mainresultstrain & \textbf{81.38} & \textbf{63.60} & \textbf{50.62} & \textbf{83.24} & \textbf{86.75} & \textbf{76.86} & \textbf{76.38} & \textbf{88.39} & \textbf{70.14} & \textbf{81.96} & \textbf{73.20} & \textbf{93.75} & \textbf{99.60\%} \scriptsize{\textbf{\textcolor{blue!90!black}{(-0.40\%)}}} \\
\midrule

\multicolumn{14}{c}{\cellcolor[HTML]{EFF8FC}\textit{Avg. Retention $\sim$12.05\% Visual Tokens}} \\
DivPrune \scriptsize{(CVPR'25)} \mainresultsfree & 75.77 & 48.46 & 45.21 & 75.91 & 84.71 & 67.18 & 73.88 & 82.84 & 61.09 & 71.72 & 40.90 & 56.43 & 84.75\% \scriptsize{\textcolor{gray}{(-15.25\%)}} \\
ZOO-Prune \scriptsize{(CVPR'26)} \mainresultsfree & 73.61 & 50.33 & 45.07 & 79.58 & 84.26 & 68.23 & 71.88 & 84.76 & 59.73 & 68.91 & 26.90 & 50.01 & 82.81\% \scriptsize{\textcolor{gray}{(-17.19\%)}} \\
EntropyPrune \scriptsize{(EMNLP'26)} \mainresultsfree & 74.13 & 47.86 & 44.64 & 75.87 & 83.16 & 66.22 & 70.24 & 84.86 & 59.04 & 71.85 & 33.60 & 51.97 & 82.60\% \scriptsize{\textcolor{gray}{(-17.40\%)}} \\
STAR-Pro \scriptsize{(Arxiv'2609)} \mainresultsfree & 77.63 & 55.73 & 44.57 & 80.15 & 85.33 & 70.84 & 73.00 & 88.12 & 63.72 & 75.06 & 32.30 & 64.49 & 87.61\% \scriptsize{\textcolor{gray}{(-12.39\%)}} \\
\noalign{\vskip\aboverulesep}
\hdashline
\noalign{\vskip\belowrulesep}
SCOPD+ \scriptsize{(Arxiv'2609)} \mainresultstrain & 80.34 & 60.40 & 47.14 & 68.06 & 85.89 & 69.54 & 71.37 & 87.84 & 63.82 & 79.97 & 43.10 & 70.43 & 89.48\% \scriptsize{\textcolor{gray}{(-10.52\%)}} \\
LT-OPD \scriptsize{(Arxiv'2609)} \mainresultstrain & 79.56 & 57.40 & 46.86 & 76.43 & 84.80 & 73.85 & 75.00 & 86.59 & 65.50 & 79.53 & 61.60 & 72.60 & 92.62\% \scriptsize{\textcolor{gray}{(-7.38\%)}} \\
\rowcolor[HTML]{FFF9D9} \textbf{SA-MoD (Ours)} \mainresultstrain & \textbf{80.84} & \textbf{61.86} & \textbf{49.98} & \textbf{80.63} & \textbf{85.77} & \textbf{74.83} & \textbf{76.88} & \textbf{88.40} & \textbf{68.73} & \textbf{81.56} & \textbf{62.60} & \textbf{91.71} & \textbf{97.25\%} \scriptsize{\textbf{\textcolor{blue!90!black}{(-2.75\%)}}} \\
\midrule

\multicolumn{14}{c}{\cellcolor[HTML]{EFF8FC}\textit{Avg. Retention $\sim$5.18\% Visual Tokens}} \\
DivPrune \scriptsize{(CVPR'25)} \mainresultsfree & 65.24 & 41.80 & 42.41 & 68.59 & 75.77 & 61.69 & 70.38 & 65.63 & 58.14 & 63.63 & 28.70 & 33.91 & 73.75\% \scriptsize{\textcolor{gray}{(-26.25\%)}} \\
ZOO-Prune \scriptsize{(CVPR'26)} \mainresultsfree & 60.52 & 40.53 & 42.04 & 73.82 & 76.89 & 63.13 & 68.63 & 71.69 & 58.04 & 63.02 & 9.40 & 28.75 & 71.82\% \scriptsize{\textcolor{gray}{(-28.18\%)}} \\
EntropyPrune \scriptsize{(EMNLP'26)} \mainresultsfree & 64.87 & 41.03 & 41.90 & 71.56 & 77.96 & 64.05 & 67.88 & 69.87 & 59.02 & 64.39 & 19.40 & 30.19 & 73.37\% \scriptsize{\textcolor{gray}{(-26.63\%)}} \\
STAR-Pro \scriptsize{(Arxiv'2609)} \mainresultsfree & 68.65 & 47.93 & 44.57 & 78.58 & 82.62 & 67.97 & 70.75 & 83.96 & 58.56 & 64.89 & 13.40 & 37.86 & 78.49\% \scriptsize{\textcolor{gray}{(-21.51\%)}} \\
\noalign{\vskip\aboverulesep}
\hdashline
\noalign{\vskip\belowrulesep}
SCOPD+ \scriptsize{(Arxiv'2609)} \mainresultstrain & 77.01 & 54.46 & 47.32 & 53.92 & 82.79 & 64.96 & 64.75 & 87.46 & 59.72 & 74.06 & 29.90 & 49.33 & 81.13\% \scriptsize{\textcolor{gray}{(-18.87\%)}} \\
LT-OPD \scriptsize{(Arxiv'2609)} \mainresultstrain & 79.95 & 57.80 & 47.27 & 75.29 & 83.68 & 71.11 & 73.50 & 86.17 & 62.46 & 77.72 & 56.40 & 59.25 & 89.84\% \scriptsize{\textcolor{gray}{(-10.16\%)}} \\
\rowcolor[HTML]{FFF9D9} \textbf{SA-MoD (Ours)} \mainresultstrain & \textbf{80.49} & \textbf{60.73} & \textbf{49.50} & \textbf{78.53} & \textbf{84.83} & \textbf{73.33} & \textbf{75.88} & \textbf{88.36} & \textbf{67.60} & \textbf{81.28} & \textbf{57.20} & \textbf{90.09} & \textbf{95.57\%} \scriptsize{\textbf{\textcolor{blue!90!black}{(-4.43\%)}}} \\
\bottomrule
\end{tabular}%
\end{adjustbox}
\end{table}

%% file: figs/amod_sparsity.tex
\begin{figure}[t]
  \centering
  \begin{minipage}[c]{0.37\linewidth}
    \centering
    \includegraphics[width=\linewidth]{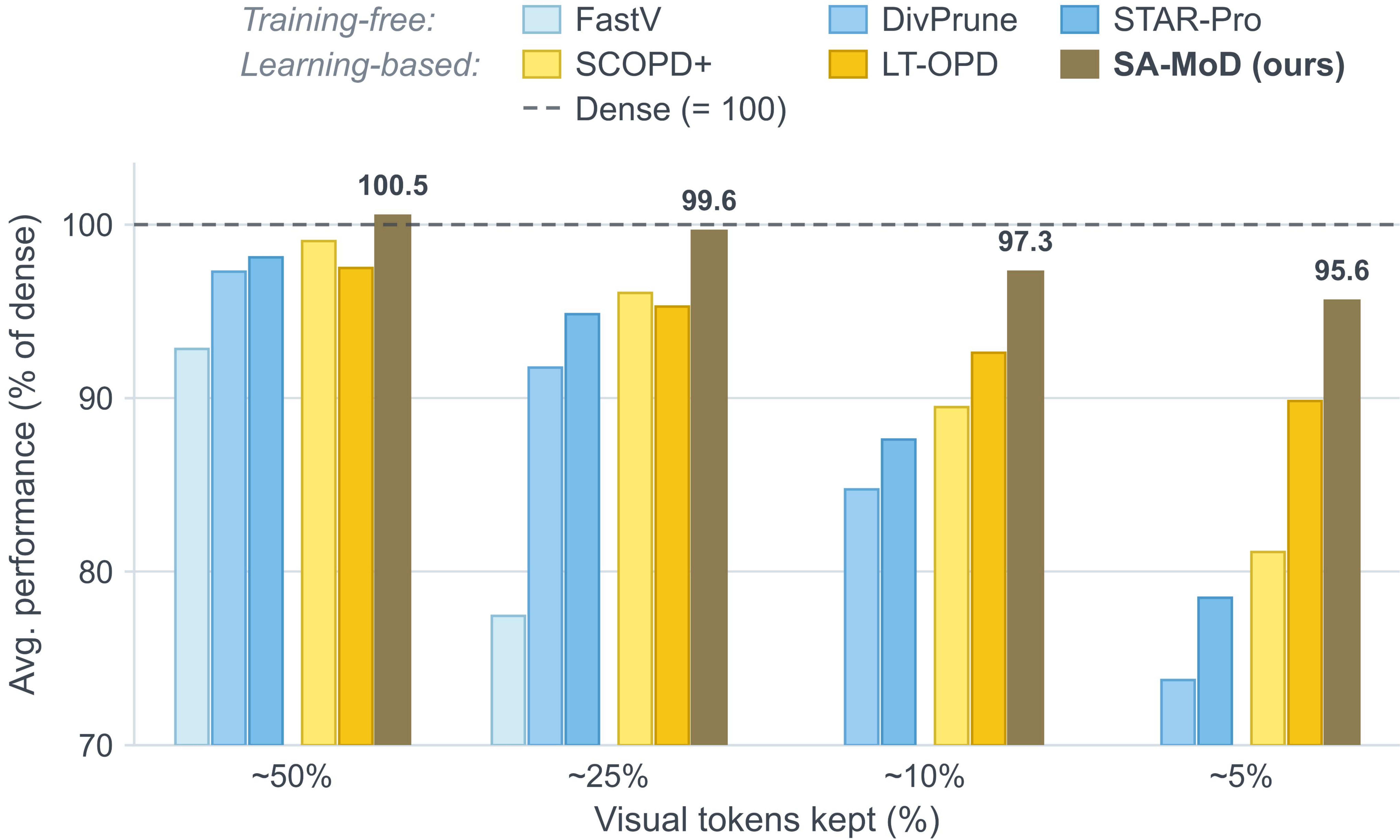}
    \vspace{-6mm}
    \caption{\textbf{Robustness across four visual token retention levels.} As retention drops from 50\% to 5\%, competing baselines degrade sharply, whereas SA-MoD drops only slightly (100.5\% to 95.6\%).}
    \label{fig:robustness}
  \end{minipage}
  \hfill
  \begin{minipage}[c]{0.60\linewidth}
    \captionof{table}{\textbf{SA-MoD across model scales and VLM architectures.} With the target retention rate $\hat{R}=25\%$, SA-MoD retains $99.38\%$ of dense performance on Qwen3.5-9B, $98.64\%$ on Qwen3-VL-4B-Instruct, and $98.26\%$ on LLaVA-OneVision-1.5-4B across the 8-benchmark suite (shown only 3 representative benchmarks). The full per-benchmark breakdown is provided in Appendix~\cref{tab:supp_cross_architecture}.}
    \label{tab:scales_arch}
    \centering
    \renewcommand{\arraystretch}{1.05}
    \resizebox{\linewidth}{!}{%
      \begin{tabular}{lcccc}
        \toprule
        \textbf{Architecture} & \textbf{MMB$_{\text{V1.1}}$} & \textbf{HR-4K} & \textbf{DocVQA} & \textbf{Avg. Perf.$^8$ (\%)} \\
        \midrule
        Qwen3.5-9B & 85.60 & 78.75 & 95.40 & 100\% \\
        + SA-MoD ($\bar{R} = 25.10\%$) & \textbf{84.40} & \textbf{79.88} & \textbf{93.54} & \textbf{99.38\%} \\
        \midrule
        Qwen3-VL-4B-Instruct & 84.21 & 76.12 & 94.79 & 100\% \\
        + SA-MoD ($\bar{R} = 24.75\%$) & \textbf{82.39} & \textbf{76.33} & \textbf{92.02} & \textbf{98.64\%} \\
        \midrule
        LLaVA-OneVision-1.5-4B & 82.89 & 58.50 & 97.83 & 100\% \\
        + SA-MoD ($\bar{R} = 24.46\%$) & \textbf{80.92} & \textbf{58.68} & \textbf{95.82} & \textbf{98.26\%} \\
        \bottomrule
      \end{tabular}%
    }
  \end{minipage}
\end{figure}

%% file: tabs/video_results.tex
\begin{table}[htbp]
\centering
\caption{\textbf{Comparison of different video token reduction methods on Qwen3.5-4B across five video benchmarks at various token retention rates.}
\mainresultsfree and \mainresultstrain indicate training-free methods and learning-based methods, respectively. 
}
\label{tab:video_results}
\setlength{\tabcolsep}{6pt}
\begin{adjustbox}{max width=\linewidth}
\begin{tabular}{lccccccccc}
\toprule
\multirow{2}{*}{\textbf{Method}} & \multirow{2}{*}{\textbf{Video-MME}} & \multirow{2}{*}{\textbf{MVBench}} & \multirow{2}{*}{\textbf{MLVU$_{\textbf{dev}}$}} & \multirow{2}{*}{\textbf{LongVideoBench$_{\text{w/o sub}}$}} & \multicolumn{4}{c}{\textbf{OmniCompass}} & \multirow{2}{*}{\textbf{Avg. Perf. (\%)}} \\
\cmidrule(lr){6-9}
 & & & & & \textbf{CS$^{\textbf{qa}}$} & \textbf{CS$^{\textbf{sg}}$} & \textbf{CS$^{\textbf{tg}}$} & \textbf{CS$^{\textbf{avg}}$} & \\
\midrule
\multicolumn{10}{c}{\cellcolor[HTML]{EFF8FC}\textit{Dense Base Model (100\% Retention)}} \\
Qwen3.5-4B & 68.62 & 67.71 & 70.43 & 64.77 & 49.75 & 12.87 & 10.93 & 24.52 & 100\% \\
\midrule

\multicolumn{10}{c}{\cellcolor[HTML]{EFF8FC}\textit{Avg. Retention $\sim$25.89\% Visual Tokens}} \\
DyCoke \scriptsize{(CVPR'25)} \mainresultsfree           & 66.86 & 64.62 & 61.25 & 58.73 & 42.31 & 1.41 & 6.75 & 16.82 & 87.82\% \scriptsize{\textcolor{gray}{(-12.18\%)}} \\
VisionZip \scriptsize{(CVPR'25)} \mainresultsfree        & 63.88 & 64.65 & 68.35 & 61.85 & 44.14 & 1.03 & 7.23 & 17.47 & 90.47\% \scriptsize{\textcolor{gray}{(-9.53\%)}} \\
HoliTom \scriptsize{(NeurIPS'25)} \mainresultsfree       & 63.40 & 65.34 & 68.80 & 62.68 & 44.02 & 1.15 & 8.20 & 17.79 & 91.18\% \scriptsize{\textcolor{gray}{(-8.82\%)}} \\
FastVID \scriptsize{(NeurIPS'25)} \mainresultsfree       & 63.92 & 64.92 & 68.71 & 62.00 & 44.63 & 0.70 & 5.60 & 17.02 & 90.34\% \scriptsize{\textcolor{gray}{(-9.66\%)}} \\
VidCom$^2$ \scriptsize{(EMNLP'25)} \mainresultsfree      & 67.44 & 65.36 & 68.29 & 62.67 & 42.43 & 0.77 & 7.54 & 16.92 & 91.51\% \scriptsize{\textcolor{gray}{(-8.49\%)}} \\
V-CAST \scriptsize{(NeurIPS'26)} \mainresultsfree        & 66.22 & 64.94 & 68.40 & 63.42 & 43.29 & 0.84 & 7.68 & 17.27 & 91.58\% \scriptsize{\textcolor{gray}{(-8.42\%)}} \\
\noalign{\vskip\aboverulesep}
\hdashline
\noalign{\vskip\belowrulesep}
SCOPD+ \scriptsize{(Arxiv'2609)} \mainresultstrain       & 67.45 & 66.89 & 69.65 & 64.70 & 43.85 & 2.07 & 8.75 & 18.22 & 94.04\% \scriptsize{\textcolor{gray}{(-5.96\%)}} \\
LT-OPD \scriptsize{(Arxiv'2609)} \mainresultstrain       & 67.00 & 65.84 & 69.67 & 64.42 & 43.98 & 2.40 & 8.15 & 18.18 & 93.48\% \scriptsize{\textcolor{gray}{(-6.52\%)}} \\
\rowcolor[HTML]{FFF9D9} \textbf{SA-MoD (Ours)} \mainresultstrain & \textbf{68.33} & \textbf{68.48} & \textbf{71.17} & \textbf{66.09} & \textbf{47.43} & \textbf{5.34} & \textbf{9.81} & \textbf{20.86} & \textbf{97.78\%} \scriptsize{\textbf{\textcolor{blue!90!black}{(-2.22\%)}}} \\
\midrule

\multicolumn{10}{c}{\cellcolor[HTML]{EFF8FC}\textit{Avg. Retention $\sim$10.76\% Visual Tokens}} \\
VisionZip \scriptsize{(CVPR'25)} \mainresultsfree        & 60.81 & 60.92 & 65.71 & 60.65 & 37.80 & 1.02 & 5.78 & 14.86 & 85.23\% \scriptsize{\textcolor{gray}{(-14.77\%)}} \\
HoliTom \scriptsize{(NeurIPS'25)} \mainresultsfree       & 58.78 & 61.65 & 65.28 & 60.80 & 40.60 & 1.13 & 7.18 & 16.31 & 85.96\% \scriptsize{\textcolor{gray}{(-14.04\%)}} \\
FastVID \scriptsize{(NeurIPS'25)} \mainresultsfree       & 60.44 & 62.52 & 66.37 & 59.76 & 40.60 & 0.69 & 5.89 & 15.73 & 86.21\% \scriptsize{\textcolor{gray}{(-13.79\%)}} \\
VidCom$^2$ \scriptsize{(EMNLP'25)} \mainresultsfree      & 63.88 & 60.89 & 63.96 & 60.36 & 37.07 & 0.49 & 4.28 & 13.95 & 84.78\% \scriptsize{\textcolor{gray}{(-15.22\%)}} \\
V-CAST \scriptsize{(NeurIPS'26)} \mainresultsfree        & 60.74 & 60.71 & 65.02 & 61.10 & 39.63 & 0.83 & 5.98 & 15.48 & 85.59\% \scriptsize{\textcolor{gray}{(-14.41\%)}} \\
\noalign{\vskip\aboverulesep}
\hdashline
\noalign{\vskip\belowrulesep}
SCOPD+ \scriptsize{(Arxiv'2609)} \mainresultstrain       & 64.88 & 65.61 & 68.89 & 62.26 & 39.79 & 1.93 & 5.91 & 15.88 & 90.03\% \scriptsize{\textcolor{gray}{(-9.97\%)}} \\
LT-OPD \scriptsize{(Arxiv'2609)} \mainresultstrain       & 65.16 & 65.44 & 69.04 & 62.85 & 40.23 & 2.13 & 5.79 & 16.05 & 90.42\% \scriptsize{\textcolor{gray}{(-9.58\%)}} \\
\rowcolor[HTML]{FFF9D9} \textbf{SA-MoD (Ours)} \mainresultstrain & \textbf{67.59} & \textbf{68.10} & \textbf{71.60} & \textbf{66.34} & \textbf{46.09} & \textbf{4.17} & \textbf{9.07} & \textbf{19.78} & \textbf{96.77\%} \scriptsize{\textbf{\textcolor{blue!90!black}{(-3.23\%)}}} \\
\bottomrule
\end{tabular}%
\end{adjustbox}
\end{table}

%% file: tabs/main_efficiency.tex
\begin{table}[htbp]
\centering
\vspace{+3mm}
\caption{\textbf{Efficiency comparison of SA-MoD  under different video context lengths.} Evaluation is conducted on \textbf{Qwen3.5-4B} with \textbf{LongVideoBench$_{\text{w/o sub}}$}, with different video processing parameters. $F_{\text{max}}$ and fps represent max number of frames and frames-per-second, respectively. 
For each setting, $\Delta$ and \textit{Speedup} denote the reduction percentage and time speedup relative to the base model (100\%).
}
\label{tab:main_efficiency}
\begin{adjustbox}{max width=\textwidth}
\begin{tabular}{lcccccccccc}
\toprule
\multirow{2}{*}{\textbf{Settings}} & \multirow{2}{*}{\shortstack{\textbf{Retention}}} & \multicolumn{2}{c}{\textbf{Latency (s)} $\downarrow$} & \multicolumn{2}{c}{\textbf{TTFT (s)} $\downarrow$} & \multicolumn{2}{c}{\textbf{TFLOPs} $\downarrow$} & \multicolumn{2}{c}{\textbf{KV-Cache (MB)} $\downarrow$} & \multirow{2}{*}{\textbf{Accuracy $\uparrow$}} \\
\cmidrule(lr){3-4} \cmidrule(lr){5-6} \cmidrule(lr){7-8} \cmidrule(lr){9-10}
& & \textbf{Value} & \textbf{Speedup} $\uparrow$ & \textbf{Value} & \textbf{Speedup} $\uparrow$ & \textbf{Value} & \textbf{$\Delta$} & \textbf{Value} & \textbf{$\Delta$} & \\
\midrule
\multirow{3}{*}{\shortstack[c]{$F_{\text{max}}=512$\\\scriptsize{(fps = 2.0)}}} & 100\% & 35.78 & 1.00$\times$ & 35.64 & 1.00$\times$ & 4872.81 & -0.0\% & 4344.99 & -0.0\% & 65.29 \\
& 27.67\% & 15.34 & \textbf{2.33$\times$} & 15.25 & \textbf{2.34$\times$} & 1492.11 & \textbf{-69.4\%} & 1438.02 & \textbf{-66.9\%} & \textbf{67.39} \\
& 11.90\% & 9.06 & \textbf{3.95$\times$} & 8.98 & \textbf{3.97$\times$} & 890.55 & \textbf{-81.7\%} & 631.84 & \textbf{-85.5\%} & \textbf{67.69} \\
\noalign{\vskip\aboverulesep}
\hdashline
\noalign{\vskip\belowrulesep}
\multirow{3}{*}{\shortstack[c]{$F_{\text{max}}=256$\\\scriptsize{(fps = 2.0)}}} & 100\% & 13.78 & 1.00$\times$ & 13.46 & 1.00$\times$ & 1689.37 & -0.0\% & 2276.55 & -0.0\% & 64.62 \\
& 28.13\% & 6.07 & \textbf{2.27$\times$} & 5.99 & \textbf{2.25$\times$} & 639.35 & \textbf{-62.2\%} & 762.09 & \textbf{-66.5\%} & \textbf{67.12} \\
& 12.54\% & 4.64 & \textbf{2.97$\times$} & 4.58 & \textbf{2.94$\times$} & 433.86 & \textbf{-74.3\%} & 339.67 & \textbf{-85.1\%} & \textbf{66.94} \\
\noalign{\vskip\aboverulesep}
\hdashline
\noalign{\vskip\belowrulesep}
\multirow{3}{*}{\shortstack[c]{$F_{\text{max}}=128$\\\scriptsize{(fps = 1.0)}}} & 100\% & 5.58 & 1.00$\times$ & 5.50 & 1.00$\times$ & 631.60 & -0.0\% & 1137.23 & -0.0\% & 63.50 \\
& 28.68\% & 2.97 & \textbf{1.88$\times$} & 2.92 & \textbf{1.88$\times$} & 280.50 & \textbf{-55.6\%} & 381.62 & \textbf{-66.4\%} & \textbf{64.25} \\
& 13.06\% & 2.37 & \textbf{2.35$\times$} & 2.32 & \textbf{2.37$\times$} & 210.24 & \textbf{-66.7\%} & 176.93 & \textbf{-84.4\%} & \textbf{65.07} \\
\bottomrule
\end{tabular}
\end{adjustbox}
\end{table}

%% file: tabs/ablation_panel.tex
\begin{table}[t]
\centering
\begin{minipage}[t]{0.535\linewidth}
  \centering
  \caption{\textbf{Ablation on progressive sparsification (PRS) and content adaptivity (CA).} Under $p$-MoD, \textit{TopK-25\%} enforces a fixed budget, \textit{TopP-0.5} enables content-adaptive routing via fixed probability-mass filtering, and \textit{CTopK-25\%} applies a prescribed curriculum schedule with fixed budgets. Full results are in the Appendix \cref{tab:supp_ablation_prs_ca}.}
  \label{tab:ablation_prs_ca}
  \vspace{-4pt}
  \setlength{\tabcolsep}{7pt}
  \renewcommand{\arraystretch}{1.08}
  \resizebox{\linewidth}{!}{%
    \begin{tabular}{lcccc}
    \toprule
    \textbf{Training Settings} & \textbf{PRS} & \textbf{CA} & \textbf{Retention} & \textbf{Avg. Perf.$^\text{8}$ (\%)} \\
    \midrule
    \rowcolor[HTML]{EFF8FC} Qwen3.5-4B & -- & -- & 100\% & 100\% \\
    \midrule
    $p$-MoD TopK-25\% & $\times$ & $\times$ & 25.00\% & 90.74\% \scriptsize{(-9.26\%)} \\
    $p$-MoD TopP-0.5 & $\times$ & \checkmark & 26.79\% & 89.61\% \scriptsize{(-10.39\%)} \\
    $p$-MoD CTopK-25\% & \checkmark & $\times$ & 25.00\% & 97.54\% \scriptsize{(-2.46\%)} \\
    \rowcolor[HTML]{FFF9D9} \textbf{SA-MoD (Ours)} & \checkmark & \checkmark & \textbf{25.06\%} & \textbf{100.26\%} \scriptsize{\textbf{\textcolor{blue!90!black}{(+0.26\%)}}} \\
    \bottomrule
    \end{tabular}%
  }
\end{minipage}
\hfill
\begin{minipage}[t]{0.435\linewidth}
  \centering
  \caption{\textbf{Ablation on SA-MoD router configurations.} \textit{w/o Param.Thresholding} uses a fixed threshold, \textit{w/o Density-aware Coef.} drops the density-aware coefficient, and \textit{w/o RouterNorm} removes RouterNorm. Full results can be found in the Appendix \cref{tab:supp_ablation_amod_router}.}
  \label{tab:ablation_amod_router}
  \vspace{-2pt}
  \setlength{\tabcolsep}{3.5pt}
  \renewcommand{\arraystretch}{1.08}
  \resizebox{\linewidth}{!}{%
    \begin{tabular}{lcc}
    \toprule
    \textbf{Training Settings} & \textbf{Retention} & \textbf{Avg. Perf.$^\text{8}$ (\%)} \\
    \midrule
    \rowcolor[HTML]{EFF8FC} Qwen3.5-4B & 100\% & 100\% \\
    \midrule
    \rowcolor[HTML]{FFF9D9} \textbf{+ SA-MoD (Ours)} & \textbf{25.06\%} & \textbf{100.26\%} \scriptsize{\textbf{\textcolor{blue!90!black}{(+0.26\%)}}} \\
    \quad w/o Param.Thresholding & 25.20\% & 91.15\% \scriptsize{(-8.85\%)} \\
    \quad w/o Density-aware Coef. & 25.12\% & 98.32\% \scriptsize{(-1.68\%)} \\
    \quad w/o RouterNorm & 25.37\% & 98.02\% \scriptsize{(-1.98\%)} \\
    \bottomrule
    \end{tabular}%
  }
\end{minipage}

\vspace{3.5mm}

\begin{minipage}[t]{0.485\linewidth}
  \centering
  \caption{\textbf{Ablation on training strategies,} including SFT, GRPO, OffPD (off-policy distillation), and OPD. Full results are in the Appendix \cref{tab:supp_ablation_training_strategy}.}
  \label{tab:ablation_training_strategy}
  \vspace{-2pt}
  \setlength{\tabcolsep}{3pt}
  \renewcommand{\arraystretch}{1.08}
  \resizebox{\linewidth}{!}{%
    \begin{tabular}{lcccc}
    \toprule
    \textbf{Training Settings} & \textbf{Retention} & \textbf{RealWQA} & \textbf{DocVQA} & \textbf{Avg. Perf.$^\text{8}$ (\%)} \\
    \midrule
    \rowcolor[HTML]{EFF8FC} Qwen3.5-4B & 100\% & 75.42 & 95.24 & 100\% \\
    \midrule
    + SA-MoD SFT   & 25.70\% & 73.54 & 73.59 & 94.68\% \scriptsize{(-5.32\%)} \\
    + SA-MoD GRPO  & 25.50\% & 75.65 & 90.08 & 98.28\% \scriptsize{(-1.72\%)} \\
    + SA-MoD OffPD & 25.48\% & 75.03 & 88.97 & 96.02\% \scriptsize{(-3.98\%)} \\
    \rowcolor[HTML]{FFF9D9} \textbf{+ SA-MoD OPD} & \textbf{25.06\%} & \textbf{76.86} & \textbf{93.75} & \textbf{100.26\%} \scriptsize{\textbf{\textcolor{blue!90!black}{(+0.26\%)}}} \\
    \bottomrule
    \end{tabular}%
  }
\end{minipage}
\hfill
\begin{minipage}[t]{0.485\linewidth}
  \centering
  \caption{\textbf{Ablation on Online Hidden State Alignment (OHiSA).} \textit{w/o OHiSA} is pure OPD, and \textit{w/ naive LD} replaces OHiSA with naive layer-wise distillation~\cite{yang2026oprd}. Full results are in the Appendix \cref{tab:supp_ablation_ohisa}.}
  \label{tab:ablation_ohisa}
  \vspace{-2pt}
  \setlength{\tabcolsep}{3pt}
  \renewcommand{\arraystretch}{1.08}
  \resizebox{\linewidth}{!}{%
    \begin{tabular}{lcccc}
    \toprule
    \textbf{Training Settings} & \textbf{Retention} & \textbf{RealWQA} & \textbf{DocVQA} & \textbf{Avg. Perf.$^\text{8}$ (\%)} \\
    \midrule
    \rowcolor[HTML]{EFF8FC} Qwen3.5-4B & 100\% & 75.42 & 95.24 & 100\% \\
    \midrule
    \rowcolor[HTML]{FFF9D9} \textbf{+ SA-MoD (Ours)} & \textbf{25.06\%} & \textbf{76.86} & \textbf{93.75} & \textbf{100.26\%} \scriptsize{\textbf{\textcolor{blue!90!black}{(+0.26\%)}}} \\
    \quad w/o OHiSA & 25.11\% & 76.42 & 93.18 & 99.67\% \scriptsize{(-0.33\%)} \\
    \quad w/ naive LD & 25.09\% & 73.46 & 89.05 & 96.99\% \scriptsize{(-3.01\%)} \\
    \bottomrule
    \end{tabular}%
  }
\end{minipage}
\end{table}

%% file: sec/5_conclusion.tex
\section{Conclusion}
We introduced Sparsity-Adaptive Mixture-of-Depth (SA-MoD), a single-stage post-training framework that progressively learns content-adaptive layer-wise visual token routing for efficient vision-language models. Rather than relying on hand-crafted token budgets or multi-stage curricula, our router combines \textit{Parameterized Thresholding} with \textit{RouterNorm} to stabilize routing scores and allocate tokens dynamically according to input complexity at each layer, while \textit{Density-aware Regularization} drives smooth progressive sparsification without a predefined schedule. Furthermore, \textit{Online Hidden-State Alignment} (OHiSA) complements On-Policy Distillation (OPD) with lightweight online hidden-state supervision to preserve intermediate representations under high compression. On Qwen3.5-4B, SA-MoD preserves 95.57\% of the base model's performance across 12 image benchmarks at 5.18\% token retention and 96.77\% across 5 video benchmarks at 10.76\% retention, while delivering up to a 3.95$\times$ end-to-end speedup under long video contexts and supporting post-hoc retention adjustment at inference time.

\vspace{+1mm}
\noindent\textbf{Limitations and Future Work.}
SA-MoD is currently restricted to visual token routing, leaving text tokens dense across layers. Exploring adaptive routing across full sequences, including both text and vision tokens, remains a promising direction  for future work.

%% file: sec/A_appendix.tex
\clearpage
{
  \titleformat{name=\section,numberless}
    {\fontsize{18pt}{22pt}\selectfont\bfseries\color{vclabink}}
    {}{0pt}{}
  \section*{Appendix}
}
\vspace{+2mm}

\noindent
In this supplementary material, we provide additional implementation details, experimental results, and qualitative analyses of SA-MoD, organized as follows:
\vspace{+4.5pt}
\begin{itemize}[itemsep=3pt, topsep=2pt, parsep=3pt]
    \item \textbf{Appendix~\ref{appendix:exp_details}} details the composition of 
    the post-training dataset, training configurations, and benchmark evaluation protocols. It also describes the deployment details  covering full- vs.\ linear-attention execution and reports GPU memory profiling.
    \item \textbf{Appendix~\ref{appendix:full_bench}}  completes results across 8 benchmarks for the cross-architecture generalization and ablation studies presented in the main paper. It also investigates training-free retention adjustment via inference-time criterion-bias perturbation.
    \item \textbf{Appendix~\ref{appendix:qualitative}} visualizes content-adaptive and layer-wise routing distributions, presents complexity-driven qualitative cases, and analyzes training dynamics and loss convergence.
\end{itemize}

\section{Additional Experimental Details}
\label{appendix:exp_details}

\subsection{Dataset Composition}
\label{app:datasets}

\noindent\textbf{Post-training Dataset (\textit{SA-MoD-Image-25K} \& \textit{SA-MoD-Video-16K}).}
We construct \textit{SA-MoD-Image-25K} by sampling 25K examples across four distinct task categories from Honey-Data-15M~\cite{zhang2026bee}, including General VQA, STEM, Grounding \& Counting, and OCR \& Document.
As detailed in Tab.~\ref{tab:supp_image_dataset}, this mixture balances general multimodal reasoning (60\%) with dense, fine-grained visual tasks such as OCR and document understanding (16\%), as well as STEM and spatial grounding (12\% each), exposing the router to diverse visual complexities and tasks.
These categories stress different aspects of visual perception: STEM problems require precise encoding of mathematical diagrams~\cite{zhang2025mavis}, while grounding and region-level tasks depend on fine-grained, localized visual cues~\cite{lin2025draw,lin2025perceive,wang2024mrmllm,wang2026locateanything}.
For video domain post-training dataset, we construct \textit{SA-MoD-Video-16K} by sampling 16K examples from Video-R1-260K~\cite{feng2025videor1reinforcingvideoreasoning}, following the curation protocol of CaCoVID~\cite{ma2026contributionawaretokencompressionefficient}.

\vspace{+1mm}
\noindent\textbf{Dataset Ablation with Learning-based Baselines.} 
To ensure a fair comparison, we post-train SA-MoD and all learning-based baselines on \textit{SA-MoD-Image-25K} and \textit{SA-MoD-Video-16K}.
As reported in \cref{tab:supp_datasets_compare}, we compare our dataset with other post-training corpora on the full 12-benchmark suite. This comparison is restricted to image datasets since the baselines only provide image-based training data.
Under the same method and matched retention budget, \textit{SA-MoD-Image-25K} consistently outperforms their original corpora (from 79.19\% to 81.13\% for SCOPD+~\cite{jeddi2026scopd} and from 87.67\% to 89.84\% for LT-OPD~\cite{li2026fewer-ltopd}), confirming the quality of our curated datasets.

\input{tabs/supp_datasets_compare}

\subsection{Training Configuration}
\label{app:train_conf}

For single-stage On-Policy Distillation (OPD), we adopt the native chat template of each base VLM without introducing routing-specific prompt instructions and enable the \textit{thinking mode} for both actor rollouts and teacher supervision.
Specifically, actor rollouts are sampled with temperature $1.0$ and a maximum of $256$ new tokens. On these rollouts, the actor is  supervised against the fixed teacher using token-level JSD loss ($\alpha=0.5$, Top-$K=100$ with tail bucket).

\input{tabs/supp_hyperparam_panel}
We jointly train the vision encoder, LLM backbone, and per-layer routers at the same learning rate, enabling the vision encoder and LLM backbone to co-evolve with sparse token routing. We insert SA-MoD routers into each decoder layer of LLM backbone and initialize its weights and bias to zero, $(W_l, b_l) = \mathbf{0}$.
The constant margin is set to $\delta = 0.1$ in all layers, and the branch reweighting scale is set to $w = 0.2$ (\cref{eq:samod_reweight}). 
We enable the \textit{thinking mode} during on-policy rollouts to preserve rich, multi-step reasoning trajectories for teacher-student alignment, while following the standard evaluation with \textit{thinking mode disabled} during inference.
Each image is processed with its original aspect ratio preserved and its token count is bounded within $[64, 16384]$. 
Each video is decoded at $\mathrm{FPS} = 2.0$ with a maximum cap of $256$ frames and a maximum budget of $768$ tokens per-frame.

To obtain models at a desired retention target $\hat{R}$, SA-MoD supports two practical strategies. The first one specifies $\hat{R}$ directly and trains until retention convergence. The second one sets a low target ($e.g.$, $5\%$) and selects intermediate checkpoints at the desired rate as retention gradually decreases from $100\%$ toward $\hat{R}$. In our experiments, we adopt the latter single-run strategy to obtain multiple operating points from a single training. Specifically, checkpoints corresponding to 50\%, 25\%, 12.5\%, and 5\% retention are extracted at 500, 1,200, 1,900, and 3,400 training steps, respectively. The hyper-parameters are summarized in \cref{tab:supp_train_hyperparams}.

\subsection{Benchmarks and Evaluation Protocol}
\label{app:evaluation}

\noindent\textbf{Evaluated Benchmarks.}
We evaluate SA-MoD across a comprehensive suite of 12 image and 5 video understanding benchmarks.
For general multimodal capabilities, MMBench$_{\text{V1.1}}$~\cite{liu2024mmbench} assesses broad  perception and reasoning via bilingual multi-choice questions, MMStar~\cite{chen2024we} filters out text-solvable samples to isolate vision-indispensable reasoning, and VisOnly~\cite{kamoi2024visonlyqa} further restricts evaluation to pure visual and geometric perception without textual shortcuts. Beyond general VQA, vision-centric benchmarks probe fine-grained spatial understanding. V$^*$Bench~\cite{wu2024v} tests high-resolution visual search for small targets, CV-Bench~\cite{tong2024cambrian} evaluates 2D and 3D spatial relationships, RealWorldQA~\cite{ai2024grok} measures real-world spatial comprehension in everyday scenes, and HRBench-4K~\cite{HRBench4k} pushes perception to ultra-high 4K resolutions. To examine model reliability under visual token sparsity, POPE~\cite{li2023evaluating} probes object hallucination through balanced yes/no polling, while HallusionBench~\cite{guan2024hallusionbench} diagnoses language-driven hallucinations and visual illusions. For text- and structure-intensive tasks, AI2D~\cite{kembhavi2016diagram} focuses on science diagram parsing, OCRBench~\cite{liu2024ocrbench} covers diverse scene-text and mathematical recognition, and DocVQA~\cite{mathew2021docvqa} evaluates question answering on text-dense documents.

For video understanding, Video-MME~\cite{fu2025video} assesses multi-duration video comprehension across diverse domains, while MVBench~\cite{li2024mvbench} targets fine-grained temporal reasoning over 20 action-centric tasks. MLVU~\cite{zhou2025mlvu} and LongVideoBench~\cite{wu2024longvideobench} evaluate multi-task long-video reasoning and needle-in-a-haystack frame retrieval without subtitles. OmniCompass~\cite{liu2026omnicompass}  measures comprehensive spatio-temporal understanding in complex video scenarios; we report its official $\mathrm{CS}^{\text{avg}}$ metric.

The \textit{Avg.\ Retention} is computed by averaging per-sample retention within each benchmark, and followed by averaging the resulting values across benchmarks.

\vspace{+1mm}
\noindent\textbf{Evaluation Protocol and Preprocessing.}
Image benchmarks are evaluated using \textit{VLMEvalKit}~\cite{duan2026vlmevalkitopensourcetoolkitevaluating}, while video benchmarks follow \textit{lmms-eval}~\cite{Zhang_2025}. All evaluations use greedy decoding without thinking mode. For image inputs, the vision token count is bounded within $[256, 16384]$, with each image's original aspect ratio preserved. For video inputs, we set the maximum number of vision tokens per frame to 768. Each benchmark adopts different frame sampling rates and maximum frame caps, we summarize the exact video preprocessing configurations in \cref{tab:supp_video_eval_preproc}.

\subsection{Inference and Deployment Details}
\label{app:deployment}

\textbf{SA-MoD in Full- and Linear-Attention Layers.}
During the prefilling stage, both full-attention and linear-attention layers share the same gather-and-writeback routing flow. At each layer, given the keep-or-skip decisions for vision tokens (with all text tokens kept dense), kept tokens are physically gathered into a compact contiguous sequence before entering the layer's token-mixing and FFN blocks. 
The resulting outputs are then modulated by the branch reweighting factor and scattered back to the full sequence length.
The difference between these two layer types lies in how they retain context for auto-regressive decoding. Full-attention layers only cache kept tokens' keys-values, yielding a \textit{ragged KV-cache} whose length varies across layers. During decoding, newly generated tokens attend exclusively to this compressed, layer-specific KV-cache.
In contrast, linear-attention layers maintain a fixed-size recurrent state instead of an explicit KV-cache. Since prefilling operates solely on the gathered tokens, the resulting recurrent state naturally summarizes the reduced visual context, enabling auto-regressive decoding to proceed directly without KV-cache overhead.

\vspace{+1mm}
\noindent\textbf{Batched Inference vs. Single-request Inference.}
SA-MoD is compatible with both batched (per-GPU batch size $>1$) and single-request (per-GPU batch size $=1$) inference settings. Under batched inference, SA-MoD's content-adaptivity produces varying visual token lengths across samples at each layer.
To construct uniform rectangular tensors for batched kernels, each sample must be padded to the maximum sequence length within the batch at each layer, with attention mask applied to exclude padded positions.
However, this padding-masking strategy introduces dummy computation that diminishes efficiency speedups.
In contrast, single-request inference eliminates cross-sample padding, enabling exact physical token gathering at each layer while aligning with standard latency-sensitive long-context serving; moreover, batching long visual sequences can easily exceed available GPU memory.
Accordingly, we adopt the single-request setting consistently across both benchmark evaluations and inference efficiency profiling.

\input{tabs/supp_gpu_memory}

\vspace{+1mm}
\noindent\textbf{GPU Memory Profiling.}
While \cref{tab:main_efficiency} reports the \textit{Latency}, \textit{TTFT}, \textit{TFLOPs}, and \textit{KV-cache} savings of SA-MoD, we supplement the corresponding GPU memory analysis in \cref{tab:supp_gpu_memory}.
Under long visual context, the global peak GPU memory mostly occurs during the vision encoder's  forward pass over all sampled frames. Consequently, token reduction methods operating at or after the visual projector, including our SA-MoD, have limited influence on the peak memory.
To accurately capture memory savings during LLM prefilling and decoding stages, we additionally record the time-averaged GPU memory sampled at $0.05$s intervals throughout inference.
As shown in \cref{tab:supp_gpu_memory}, although peak memory remains nearly unchanged, SA-MoD substantially lowers \textit{the average GPU memory} across evaluated settings.

\section{Comprehensive Benchmark Results}
\label{appendix:full_bench}

\subsection{Extended Full Benchmark Results}
\label{app:full_breakdowns}

To complement the tables in the main paper, we report the complete per-benchmark results for the cross-architecture, cross-scale evaluations and ablation studies.
All experiments in this section are evaluated on a unified suite of 8 image benchmarks comprising MMBench$_{\text{V1.1}}$~\cite{liu2024mmbench}, AI2D~\cite{kembhavi2016diagram}, RealWorldQA~\cite{ai2024grok}, MMStar~\cite{chen2024we}, POPE~\cite{li2023evaluating}, CV-Bench~\cite{tong2024cambrian}, HRBench-4K~\cite{HRBench4k}, and DocVQA~\cite{mathew2021docvqa}.
Specifically, \cref{tab:supp_cross_architecture} extends \cref{tab:scales_arch} to detail the cross-architecture and cross-scale generalization on the full 8-benchmark performance.
\cref{tab:supp_ablation_prs_ca} provides the full per-benchmark breakdown of the progressive sparsification and content-adaptivity in \cref{tab:ablation_prs_ca}, while \cref{tab:supp_ablation_amod_router} details the router configuration ablations in \cref{tab:ablation_amod_router}.
Finally, \cref{tab:supp_ablation_training_strategy} and \cref{tab:supp_ablation_ohisa} present the complete per-benchmark results for the training strategy comparisons (\cref{tab:ablation_training_strategy}) and Online Hidden-States Alignment (OHiSA) ablations (\cref{tab:ablation_ohisa}), respectively.
Among the compared training strategies, GRPO~\cite{shao2024deepseekmath} serves as the reinforcement-learning alternative, which has been widely adopted for multimodal post-training~\cite{feng2025videor1reinforcingvideoreasoning,tong2025delving}.

\input{tabs/supp_cross_architecture}
\input{tabs/supp_ablation_e2e_progressive}
\input{tabs/supp_ablation_paramthresh_details}
\input{tabs/supp_ablation_training_strategy}
\input{tabs/supp_ablation_ohisd_full}

\subsection{Inference-Time Retention Adjustment}
\label{app:threshold_perturbation}

Recall that SA-MoD determines token routing decisions via the parameterized thresholding criterion in \cref{eq:routing_criterion}, where the entity-level threshold is defined as $T(H_{l,v}) = \mathrm{mean}_{j \in I_{l,v}}(t_j) + \delta$, with the margin $\delta = 0.1$ fixed during training (\cref{tab:supp_train_hyperparams}).
At inference time, a trained checkpoint supports post-hoc retention adjustment simply by adding a scalar bias $\epsilon$ exclusively to the routing criterion: $\mathbb{I}(S(h_i) + T(H_{l,v}) + \epsilon \ge 0)$. This adjustment leaves both the learned token importance ranking and symmetric branch reweighting computation completely unaffected. 
Specifically, a negative bias ($\epsilon < 0$) shifts the decision boundary downward to prune more visual tokens, whereas a positive bias serves oppositely. This provides a lightweight way to trade off accuracy and efficiency from a single checkpoint without retraining.
In practice, $\epsilon$ for any retention target can be easily calibrated via a small probing set.

To examine this capability, we take the SA-MoD checkpoint trained at $25.06\%$ retention and apply a negative criterion bias $\epsilon=-0.06$ at inference time to reach around $5\%$ retention target (\cref{tab:supp_margin_perturb}), reducing the retention by nearly $5\times$ while still preserving $97.45\%$ relative performance.
Even across benchmarks such as RealWorldQA ($73.21$ vs.\ $73.33$) and DocVQA ($89.76$ vs.\ $90.09$), the bias-adjusted model closely tracks the $5.31\%$ trained checkpoint, confirming that the router from earlier steps has already learned a well-calibrated relative token importance.
Overall, inference-time criterion-bias adjustment incurs only a $0.57\%$ gap relative to the dedicated $5.31\%$ model ($98.02\%$), enabling flexible retention adjustment without additional training.

\section{Qualitative Analysis and Case Study}
\label{appendix:qualitative}

\subsection{Training Dynamics and Loss Convergence}
\label{app:train_dynamics}

\cref{fig:training_dynamics} illustrates the optimization trajectories of SA-MoD on Qwen3.5-4B with a target retention rate of $\hat{R}=5\%$.
Owing to zero-initialized routers $(W_l, b_l) = \mathbf{0}$ and positive margin $\delta = 0.1$, training starts from a lossless $100\%$ dense state where actor and teacher outputs coincide ($\mathcal{L}_{\text{OPD}} = \mathcal{L}_{\text{OHiSA}} = 0$).
As shown in \cref{fig:training_dynamics}(b), global visual token retention decays smoothly and monotonically, enabling multi-budget checkpoints from a single run.
Correspondingly, $\mathcal{L}_{\text{DReg}}$ (\cref{fig:training_dynamics}(c)) actively drives threshold decay during the main sparsification phase (steps $0$--$2{,}500$) and attenuates once retention approaches to the $5\%$ target after $3{,}000$ steps.
Meanwhile, $\mathcal{L}_{\text{OPD}}$ (\cref{fig:training_dynamics}(a)) and $\mathcal{L}_{\text{OHiSA}}$ (\cref{fig:training_dynamics}(d)) rise smoothly until saturation.

\input{tabs/supp_margin_perturb}
\input{figs/training_dynamics}

\subsection{Sparse Routing Distribution and Case Study}
\label{app:case_studies}

\cref{fig:amod_retention_dist} demonstrates SA-MoD's \textit{content-adaptivity} and \textit{layer-wise routing} at the 25\% target retention setting, while \cref{fig:vis_cases} provides qualitative examples of complexity-aware token selection at approximately $5\%$ retention. 
As shown in the left panel of \cref{fig:amod_retention_dist}, per-sample retention varies with input content, allocating more tokens to visually demanding tasks (DocVQA and MLVU) while applying more aggressive pruning to relatively simplier contexts (MVBench and RealWorldQA). 
For the layer-wise retention distribution (\cref{fig:amod_retention_dist}, right), both image ($27.5\%$ avg.) and video ($24.1\%$ avg.) domain inputs keep $>40\%$ tokens in early layers ($\text{L1}$--$\text{L8}$) before tapering to $<10\%$ in deep layers ($\text{L24}$--$\text{L32}$). Notably, full-attention layers ($\text{L4}, \text{L8}, \dots, \text{L32}$) exhibit similar retention levels to those of adjacent linear-attention layers without noticeable discrepancy.

For the case visualization, \cref{fig:vis_cases} further highlights SA-MoD's content-adaptivity across inputs of varying visual complexity. On the $5\%$ retention checkpoint, a simple icon (top) averages $2.9\%$ retention by focusing on its body and edges, leaving only sparse and evenly spaced background tokens as positional anchors; a natural dense scene (bottom) retains $5.1\%$, with kept tokens clustered near foreground objects; and a dense line-chart document (middle) retains $6.8\%$, prioritizing the plotted curve, axes, and text labels.

\clearpage
\input{figs/amod_retention_dist}
\input{figs/vis-cases}
\clearpage

%% file: tabs/supp_datasets_compare.tex
\begin{table}[b]
\centering
\caption{\textbf{Full-suite comparison of post-training datasets across learning-based baselines.} We compare SCOPD+~\cite{jeddi2026scopd} trained on its original \textit{SCOPD-LLaVA-CoT-10K}~\cite{jeddi2026scopd} corpus ($\dagger$) and LT-OPD~\cite{li2026fewer-ltopd} trained on its original \textit{LT-OPD-14K}~\cite{li2026fewer-ltopd} corpus ($\dagger$) against training both methods on our constructed \textit{SA-MoD-Image-25K} dataset across all 12 image benchmarks, all evaluated at \textbf{5.18\%} retention.}
\vspace{-4pt}
\label{tab:supp_datasets_compare}
\begingroup
\setlength{\tabcolsep}{2.5pt}
\begin{adjustbox}{max width=\linewidth}
\begin{tabular}{lccccccccccccc}
\toprule
\textbf{Method} & \textbf{MMB$_{\text{V1.1}}$} & \textbf{AI2D$_{\text{test}}$} & \textbf{RealWQA} & \textbf{MMStar} & \textbf{POPE} & \textbf{Hallusion} & \textbf{CV-Bench} & \textbf{V$^*$Bench} & \textbf{VisOnly} & \textbf{HR-4K} & \textbf{OCRB} & \textbf{DocVQA} & \textbf{Avg. Perf.$^{\text{12}}$ (\%)} \\
\midrule
Qwen3.5-4B & 82.73 & 83.45 & 75.42 & 62.06 & 87.98 & 71.50 & 86.87 & 80.10 & 47.91 & 74.63 & 85.60 & 95.24 & 100\% \\
\midrule
SCOPD+$^\dagger$        & 76.47 & 73.90 & 64.18 & 54.20 & 87.07 & 58.88 & 81.89 & 54.02 & 45.38 & 62.00 & 27.30 & 42.02 & 79.19\% \\
\rowcolor[HTML]{FFF9D9} SCOPD+                 & 77.01 & 74.06 & 64.96 & 54.46 & 87.46 & 59.72 & 82.79 & 53.92 & 47.32 & 64.75 & 29.90 & 49.33 & \textbf{81.13\%} \scriptsize{\textbf{\textcolor{blue!90!black}{(+1.94\%)}}} \\
\noalign{\vskip\aboverulesep}
\hdashline
\noalign{\vskip\belowrulesep}
LT-OPD$^\dagger$       & 78.94 & 77.13 & 70.19 & 57.60 & 85.89 & 59.62 & 82.38 & 74.80 & 46.40 & 69.25 & 55.60 & 51.80 & 87.67\% \\
\rowcolor[HTML]{FFF9D9} LT-OPD                 & 79.95 & 77.72 & 71.11 & 57.80 & 86.17 & 62.46 & 83.68 & 75.29 & 47.27 & 73.50 & 56.40 & 59.25 & \textbf{89.84\%} \scriptsize{\textbf{\textcolor{blue!90!black}{(+2.17\%)}}} \\
\bottomrule
\end{tabular}%
\end{adjustbox}
\endgroup
\vspace{+6pt}
\end{table}

%% file: tabs/supp_hyperparam_panel.tex
\begin{table}[htbp]
  \centering
  \captionsetup{skip=4pt}
  \begin{minipage}[t]{0.525\linewidth}
    \vspace{0pt}
    \centering
    \caption{\textbf{Composition of \textit{SA-MoD-Image-25K}.} Breakdown of the 25K image post-training samples drawn from Honey-Data-15M across four task categories.}
    \label{tab:supp_image_dataset}
    \begingroup
    \setlength{\tabcolsep}{6pt}
    \renewcommand{\arraystretch}{1.12}
    \resizebox{\linewidth}{!}{%
      \begin{tabular}{lcc}
      \toprule
      \textbf{Dataset Component} & \textbf{\# Samples (\%)} & \textbf{Median Resolution} \\
      \midrule
      General VQA & 15K (60\%) & $720\times 640$ \\
      STEM & 3K (12\%) & $657\times 436$ \\
      Grounding \& Counting & 3K (12\%) & $480\times 320$ \\
      OCR \& Document & 4K (16\%) & $960\times 700$ \\
      \bottomrule
      \end{tabular}%
    }
    \endgroup

    \vspace{23pt}

    \caption{\textbf{Video preprocessing settings across benchmarks.}}
    \label{tab:supp_video_eval_preproc}
    \begingroup
    \setlength{\tabcolsep}{7pt}
    \renewcommand{\arraystretch}{1.12}
    \resizebox{\linewidth}{!}{%
      \begin{tabular}{lcc}
      \toprule
      \textbf{Benchmark} & \textbf{FPS / Fixed Num} & \textbf{Max Frames} \\
      \midrule
      Video-MME & FPS 1.0 & 128 \\
      MVBench & Fixed Num & 64 \\
      OmniCompass & Fixed Num & 64 \\
      MLVU$_{\text{dev}}$ & Fixed Num & 128 \\
      LongVideoBench$_{\text{w/o sub}}$ & Fixed Num & 128 \\
      \bottomrule
      \end{tabular}%
    }
    \endgroup
  \end{minipage}
  \hfill
  \begin{minipage}[t]{0.445\linewidth}
    \vspace{0pt}
    \centering
    \caption{\textbf{Training hyperparameters for SA-MoD.} Summary of optimization, rollout distillation, and router configurations.}
    \label{tab:supp_train_hyperparams}
    \begingroup
    \setlength{\tabcolsep}{8pt}
    \renewcommand{\arraystretch}{1.16}
    \resizebox{\linewidth}{!}{%
      \begin{tabular}{lc}
      \toprule
      \textbf{Configuration} & \textbf{Setting} \\
      \midrule
      AdamW Optimizer & ($\beta_1=0.9, \beta_2=0.999$) \\
      Weight Decay & $0.01$ \\
      Learning Rate & $2.0\times 10^{-6}$ \\
      LR Schedule & Constant \\
      LR Warmup Steps & 10 \\
      Gradient Clipping & $1.0$ \\
      Global Batch Size & $32$ \\
      Per-sample Rollout & $4$ \\
      Max Rollout Length & $256$ \\
      JSD Top-$K$ Bucket & $100$ + Tail Bucket \\
      \midrule
      DReg Loss Weight ($\lambda_{\text{DReg}}$) & $1.0\times 10^{-3}$ \\
      OHiSA Loss Weight ($\lambda_{\text{OHiSA}}$) & $1.0$ \\
      Threshold Margin $\delta$ & 0.1 \\
      TanhNorm Weight $w$ & 0.2 \\
      \bottomrule
      \end{tabular}%
    }
    \endgroup
  \end{minipage}
\end{table}

%% file: tabs/supp_gpu_memory.tex
\begin{table}[htbp]
    \centering
    \captionsetup{skip=4pt}
    \caption{\textbf{GPU memory profiling across different video context lengths.} We evaluate Qwen3.5-4B on \textit{LongVideoBench}$_{\text{w/o sub}}$ (BF16, $\text{per GPU batch size}=1$), reporting global peak and $0.05$s time-averaged memory. Peak memory remains nearly unchanged due to the fixed vision encoder overhead before LLM prefilling, while SA-MoD notably reduces time-averaged GPU memory.}
    \vspace{+1mm}
    \label{tab:supp_gpu_memory}
    \begingroup
    \small
    \setlength{\tabcolsep}{10pt}
    \renewcommand{\arraystretch}{1.08}
    \begin{tabular}{ccccc}
    \toprule
    \multirow{2}{*}{\textbf{Settings}} & \multirow{2}{*}{\textbf{Retention}} & \multirow{2}{*}{\textbf{Peak (GB)}} & \multicolumn{2}{c}{\textbf{Time-Avg.\ (GB)}} \\
    \cmidrule(lr){4-5}
    & & & \textbf{Value} & \textbf{$\Delta$} \\
    \midrule
    \multirow{3}{*}{\shortstack[c]{$F_{\max}=512$\\\scriptsize{($\text{fps}=2.0$)}}} & 100\% & 66.145 & 52.047 & -- \\
    & 27.67\% & 66.029 & 39.354 & \textbf{-24.4\%} \\
    & 11.90\% & 66.029 & 34.935 & \textbf{-32.9\%} \\
    \noalign{\vskip\aboverulesep}
    \hdashline
    \noalign{\vskip\belowrulesep}
    \multirow{3}{*}{\shortstack[c]{$F_{\max}=256$\\\scriptsize{($\text{fps}=2.0$)}}} & 100\% & 38.426 & 28.691 & -- \\
    & 28.13\% & 38.395 & 24.725 & \textbf{-13.8\%} \\
    & 12.54\% & 38.395 & 22.853 & \textbf{-20.3\%} \\
    \noalign{\vskip\aboverulesep}
    \hdashline
    \noalign{\vskip\belowrulesep}
    \multirow{3}{*}{\shortstack[c]{$F_{\max}=128$\\\scriptsize{($\text{fps}=1.0$)}}} & 100\% & 23.457 & 17.570 & -- \\
    & 28.68\% & 23.244 & 16.280 & \textbf{-7.3\%} \\
    & 13.06\% & 23.232 & 15.832 & \textbf{-9.9\%} \\
    \bottomrule
    \end{tabular}
    \endgroup
\end{table}

%% file: tabs/supp_cross_architecture.tex
\begin{table}[htbp]
\centering
\caption{\textbf{Full 8-benchmark performance of SA-MoD adaptation across scales and VLM architectures.} Complete per-benchmark results on Qwen3.5-9B, Qwen3-VL-4B-Instruct, and LLaVA-OneVision-1.5-4B under a $\sim 25\%$ target retention budget.}
\label{tab:supp_cross_architecture}
\begin{adjustbox}{max width=\linewidth}
\begin{tabular}{lccccccccc}
\toprule
\textbf{Architecture} & \textbf{MMB$_{\text{V1.1}}$} & \textbf{AI2D$_{\text{test}}$} & \textbf{RealWQA} & \textbf{MMStar} & \textbf{POPE} & \textbf{CV-Bench} & \textbf{HR-4K} & \textbf{DocVQA} & \textbf{Avg. Perf.$^\text{8}$ (\%)} \\
\midrule
Qwen3.5-9B & 85.60 & 88.21 & 78.03 & 68.86 & 89.79 & 87.43 & 78.75 & 95.40 & 100\% \\
\textbf{+ SA-MoD}~\scriptsize{($\bar{R}=25.10\%$)} & \textbf{84.40} & \textbf{86.57} & \textbf{76.97} & \textbf{68.69} & \textbf{90.28} & \textbf{87.32} & \textbf{79.88} & \textbf{93.54} & \textbf{99.38\%} \\
\midrule
Qwen3-VL-4B-Instruct & 84.21 & 84.68 & 70.84 & 62.36 & 89.27 & 85.69 & 76.12 & 94.79 & 100\% \\
\textbf{+ SA-MoD}~\scriptsize{($\bar{R}=24.75\%$)} & \textbf{82.39} & \textbf{81.93} & \textbf{70.25} & \textbf{61.86} & \textbf{89.12} & \textbf{84.78} & \textbf{76.33} & \textbf{92.02} & \textbf{98.64\%} \\
\midrule
LLaVA-OneVision-1.5-4B & 82.89 & 84.87 & 70.98 & 64.93 & 87.44 & 62.67 & 58.50 & 97.83 & 100\% \\
\textbf{+ SA-MoD}~\scriptsize{($\bar{R}=24.46\%$)} & \textbf{80.92} & \textbf{81.25} & \textbf{70.62} & \textbf{61.88} & \textbf{87.25} & \textbf{62.58} & \textbf{58.68} & \textbf{95.82} & \textbf{98.26\%} \\
\bottomrule
\end{tabular}
\end{adjustbox}
\end{table}

%% file: tabs/supp_ablation_e2e_progressive.tex
\begin{table}[htbp]
\centering
\caption{\textbf{Full 8-benchmark ablation on progressive sparsification (PRS) and content adaptivity (CA).} Keeping the OPD and OHiSA objectives fixed, we compare SA-MoD with fixed top-$k$ (TopK), fixed top-$p$ (TopP), and curriculum top-$k$ (CTopK) with cosine decay starting from 100\%, all under $p$-MoD settings.}
\label{tab:supp_ablation_prs_ca}
\begingroup
\setlength{\tabcolsep}{3pt}
\begin{adjustbox}{max width=\linewidth}
\begin{tabular}{lcccccccccccc}
\toprule
\textbf{Training Settings} & \textbf{PRS} & \textbf{CA} & \textbf{Retention} & \textbf{MMB$_{\text{V1.1}}$} & \textbf{AI2D$_{\text{test}}$} & \textbf{RealWQA} & \textbf{MMStar} & \textbf{POPE} & \textbf{CV-Bench} & \textbf{HR-4K} & \textbf{DocVQA} & \textbf{Avg.\ Perf.$^\text{8}$ (\%)} \\
\midrule
\rowcolor[HTML]{EFF8FC} Qwen3.5-4B & -- & -- & 100\% & 82.73 & 83.45 & 75.42 & 62.06 & 87.98 & 86.87 & 74.63 & 95.24 & 100\% \\
\midrule
$p$-MoD TopK-25\% & $\times$ & $\times$ & 25.00\% & 78.34 & 79.86 & 70.11 & 56.53 & 85.16 & 80.81 & 65.63 & 70.19 & 90.74\% \scriptsize{(-9.26\%)} \\
$p$-MoD TopP-0.5 & $\times$ & \checkmark & 26.79\% & 78.19 & 79.08 & 69.58 & 56.16 & 84.60 & 79.67 & 65.88 & 65.46 & 89.61\% \scriptsize{(-10.39\%)} \\
$p$-MoD CTopK-25\% & \checkmark & $\times$ & 25.00\% & 80.03 & 80.51 & 74.94 & 60.86 & 87.26 & 84.98 & 75.88 & 86.70 & 97.54\% \scriptsize{(-2.46\%)} \\
\rowcolor[HTML]{FFF9D9} \textbf{SA-MoD (Ours)} & \checkmark & \checkmark & \textbf{25.06\%} & \textbf{81.38} & \textbf{81.96} & \textbf{76.86} & \textbf{63.60} & \textbf{88.39} & \textbf{86.75} & \textbf{76.38} & \textbf{93.75} & \textbf{100.26\%} \scriptsize{\textbf{\textcolor{blue!90!black}{(+0.26\%)}}} \\
\bottomrule
\end{tabular}
\end{adjustbox}
\endgroup
\end{table}

%% file: tabs/supp_ablation_paramthresh_details.tex
\begin{table}[htbp]
\centering
\caption{\textbf{Full 8-benchmark ablation on SA-MoD Router configurations.} Keeping the OPD and OHiSA training objectives fixed, \textit{w/o Param.Thresholding} means fixed thresholding mechanism, \textit{w/o Density-aware Coef.} uses regularization loss without density-aware coefficient, and \textit{w/o RouterNorm} removes the RouterNorm module.}
\label{tab:supp_ablation_amod_router}
\begingroup
\setlength{\tabcolsep}{3pt}
\begin{adjustbox}{max width=\linewidth}
\begin{tabular}{lcccccccccc}
\toprule
\textbf{Training Settings} & \textbf{Retention} & \textbf{MMB$_{\text{V1.1}}$} & \textbf{AI2D$_{\text{test}}$} & \textbf{RealWQA} & \textbf{MMStar} & \textbf{POPE} & \textbf{CV-Bench} & \textbf{HR-4K} & \textbf{DocVQA} & \textbf{Avg.\ Perf.$^\text{8}$ (\%)} \\
\midrule
\rowcolor[HTML]{EFF8FC} Qwen3.5-4B & 100\% & 82.73 & 83.45 & 75.42 & 62.06 & 87.98 & 86.87 & 74.63 & 95.24 & 100\% \\
\midrule
\rowcolor[HTML]{FFF9D9} \textbf{+ SA-MoD (Ours)} & \textbf{25.06\%} & \textbf{81.38} & \textbf{81.96} & \textbf{76.86} & \textbf{63.60} & \textbf{88.39} & \textbf{86.75} & \textbf{76.38} & \textbf{93.75} & \textbf{100.26\%} \scriptsize{\textbf{\textcolor{blue!90!black}{(+0.26\%)}}} \\
\quad w/o Param.Thresholding & 25.20\% & 75.95 & 72.19 & 68.60 & 54.70 & 84.08 & 79.95 & 71.33 & 84.36 & 91.15\% \scriptsize{(-8.85\%)} \\
\quad w/o Density-aware Coef. & 25.12\% & 79.89 & 80.14 & 75.33 & 61.65 & 87.88 & 85.82 & 76.13 & 89.59 & 98.32\% \scriptsize{(-1.68\%)} \\
\quad w/o RouterNorm & 25.37\% & 80.14 & 79.95 & 74.65 & 62.02 & 86.95 & 85.43 & 76.33 & 88.71 & 98.02\% \scriptsize{(-1.98\%)} \\
\bottomrule
\end{tabular}
\end{adjustbox}
\endgroup
\end{table}

%% file: tabs/supp_ablation_training_strategy.tex
\begin{table}[htbp]
\centering
\caption{\textbf{Full 8-benchmark ablation on different training strategies.} Keeping the SA-MoD mechanism and OHiSA fixed, we vary the training objective including SFT, GRPO, OffPD (off-policy distillation), and OPD.}
\label{tab:supp_ablation_training_strategy}
\begingroup
\setlength{\tabcolsep}{4.5pt}
\begin{adjustbox}{max width=\linewidth}
\begin{tabular}{lcccccccccc}
\toprule
\textbf{Training Settings} & \textbf{Retention} & \textbf{MMB$_{\text{V1.1}}$} & \textbf{AI2D$_{\text{test}}$} & \textbf{RealWQA} & \textbf{MMStar} & \textbf{POPE} & \textbf{CV-Bench} & \textbf{HR-4K} & \textbf{DocVQA} & \textbf{Avg.\ Perf.$^\text{8}$ (\%)} \\
\midrule
\rowcolor[HTML]{EFF8FC} Qwen3.5-4B & 100\% & 82.73 & 83.45 & 75.42 & 62.06 & 87.98 & 86.87 & 74.63 & 95.24 & 100\% \\
\midrule
+ SA-MoD SFT & 25.70\% & 82.31 & 81.34 & 73.54 & 60.91 & 87.30 & 82.89 & 69.34 & 73.59 & 94.68\% \scriptsize{(-5.32\%)} \\
+ SA-MoD GRPO & 25.50\% & 80.29 & 80.43 & 75.65 & 62.12 & 86.89 & 84.54 & 75.92 & 90.08 & 98.28\% \scriptsize{(-1.72\%)} \\
+ SA-MoD OffPD & 25.48\% & 77.41 & 75.68 & 75.03 & 61.03 & 83.76 & 82.85 & 76.17 & 88.97 & 96.02\% \scriptsize{(-3.98\%)} \\
\rowcolor[HTML]{FFF9D9} \textbf{+ SA-MoD OPD} & \textbf{25.06\%} & \textbf{81.38} & \textbf{81.96} & \textbf{76.86} & \textbf{63.60} & \textbf{88.39} & \textbf{86.75} & \textbf{76.38} & \textbf{93.75} & \textbf{100.26\%} \scriptsize{\textbf{\textcolor{blue!90!black}{(+0.26\%)}}} \\
\bottomrule
\end{tabular}
\end{adjustbox}
\endgroup
\end{table}

%% file: tabs/supp_ablation_ohisd_full.tex
\begin{table}[htbp]
\centering
\caption{\textbf{Full 8-benchmark ablation on Online Hidden State Alignment (OHiSA).} With the SA-MoD and OPD objectives fixed, \textit{w/o OHiSA} indicates pure OPD objective without OHiSA, and \textit{w/ naive LD} means replacing OHiSA with naive layer-wise distillation~\cite{yang2026oprd}.}
\label{tab:supp_ablation_ohisa}
\begingroup
\setlength{\tabcolsep}{4.5pt}
\begin{adjustbox}{max width=\linewidth}
\begin{tabular}{lcccccccccc}
\toprule
\textbf{Training Settings} & \textbf{Retention} & \textbf{MMB$_{\text{V1.1}}$} & \textbf{AI2D$_{\text{test}}$} & \textbf{RealWQA} & \textbf{MMStar} & \textbf{POPE} & \textbf{CV-Bench} & \textbf{HR-4K} & \textbf{DocVQA} & \textbf{Avg.\ Perf.$^\text{8}$ (\%)} \\
\midrule
\rowcolor[HTML]{EFF8FC} Qwen3.5-4B & 100\% & 82.73 & 83.45 & 75.42 & 62.06 & 87.98 & 86.87 & 74.63 & 95.24 & 100\% \\
\midrule
\rowcolor[HTML]{FFF9D9} \textbf{+ SA-MoD (Ours)} & \textbf{25.06\%} & \textbf{81.38} & \textbf{81.96} & \textbf{76.86} & \textbf{63.60} & \textbf{88.39} & \textbf{86.75} & \textbf{76.38} & \textbf{93.75} & \textbf{100.26\%} \scriptsize{\textbf{\textcolor{blue!90!black}{(+0.26\%)}}} \\
\quad w/o OHiSA & 25.11\% & 80.63 & 81.70 & 76.42 & 63.47 & 87.62 & 85.83 & 76.22 & 93.18 & 99.67\% \scriptsize{(-0.33\%)} \\
\quad w/ naive LD & 25.09\% & 79.87 & 80.20 & 73.46 & 61.34 & 86.80 & 83.39 & 73.79 & 89.05 & 96.99\% \scriptsize{(-3.01\%)} \\
\bottomrule
\end{tabular}
\end{adjustbox}
\endgroup
\end{table}

%% file: tabs/supp_margin_perturb.tex
\begin{table}[t]
\centering
\captionsetup{skip=4pt}
\renewcommand{\thetable}{B6} 
\addtocounter{table}{-1}
\caption{\textbf{Inference-time retention adjustment via criterion-bias perturbation.} Using the SA-MoD checkpoint trained at $\bar{R}=25.06\%$ ($\epsilon=0$), we apply a negative criterion bias ($\epsilon < 0$) solely to the routing decision criterion at inference time without retraining or altering branch reweighting (${\ddagger}$). Specifically, we set $\epsilon=-0.06$ to adjust retention to 5.23\% in this experiment.}
\vspace{+0.5mm}
\label{tab:supp_margin_perturb}
\begingroup
\setlength{\tabcolsep}{3pt}
\begin{adjustbox}{max width=\textwidth}
\begin{tabular}{lcccccccccc}
\toprule
\textbf{Checkpoints} & \textbf{Retention} & \textbf{MMB$_{\text{V1.1}}$} & \textbf{AI2D$_{\text{test}}$} & \textbf{RealWQA} & \textbf{MMStar} & \textbf{POPE} & \textbf{CV-Bench} & \textbf{HR-4K} & \textbf{DocVQA} & \textbf{Avg.\ Perf.$^\text{8}$ (\%)} \\
\midrule
\rowcolor[HTML]{EFF8FC} Qwen3.5-4B & 100\% & 82.73 & 83.45 & 75.42 & 62.06 & 87.98 & 86.87 & 74.63 & 95.24 & 100\% \\
\midrule
+ SA-MoD & 25.06\% & 81.38 & 81.96 & 76.86 & 63.60 & 88.39 & 86.75 & 76.38 & 93.75 & 100.26\% \\
+ SA-MoD & 5.31\% & 80.49 & 81.28 & 73.33 & 60.73 & 88.36 & 84.83 & 75.88 & 90.09 & 98.02\% \\
\noalign{\vskip\aboverulesep}
\hdashline
\noalign{\vskip\belowrulesep}
+ SA-MoD$^{\ddagger}$ & 25.06\% $\xrightarrow{\epsilon = -0.06}$ 5.23\% & 80.15 & 80.46 & 73.21 & 59.86 & 88.17 & 84.08 & 75.75 & 89.76 & 97.45\% \\
\bottomrule
\end{tabular}
\end{adjustbox}
\endgroup
\vspace{+6mm}
\end{table}

%% file: figs/training_dynamics.tex
\begin{figure}[t]
    \centering
    \includegraphics[width=\linewidth]{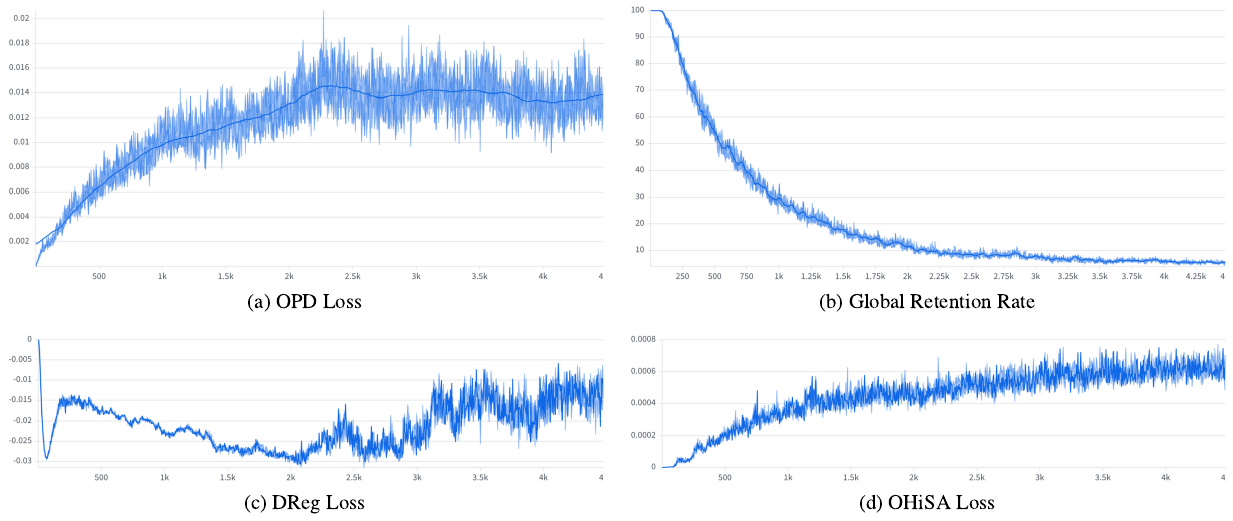}
    \caption{\textbf{Training dynamics and convergence of SA-MoD} with $\hat{R}=5\%$ on Qwen3.5-4B.
    (a) On-policy distillation loss $\mathcal{L}_{\text{OPD}}$;
    (b) Global visual token retention rate averaged across layers and samples;
    (c) Density-aware sparsity regularization loss $\mathcal{L}_{\text{DReg}}$; and
    (d) Online hidden-states alignment loss $\mathcal{L}_{\text{OHiSA}}$. The retention rate gradually decreases towards the target, while the training losses stabilize.}
    \label{fig:training_dynamics}
\end{figure}

%% file: figs/amod_retention_dist.tex
\begin{figure}[t]
    \centering
    \includegraphics[width=\linewidth]{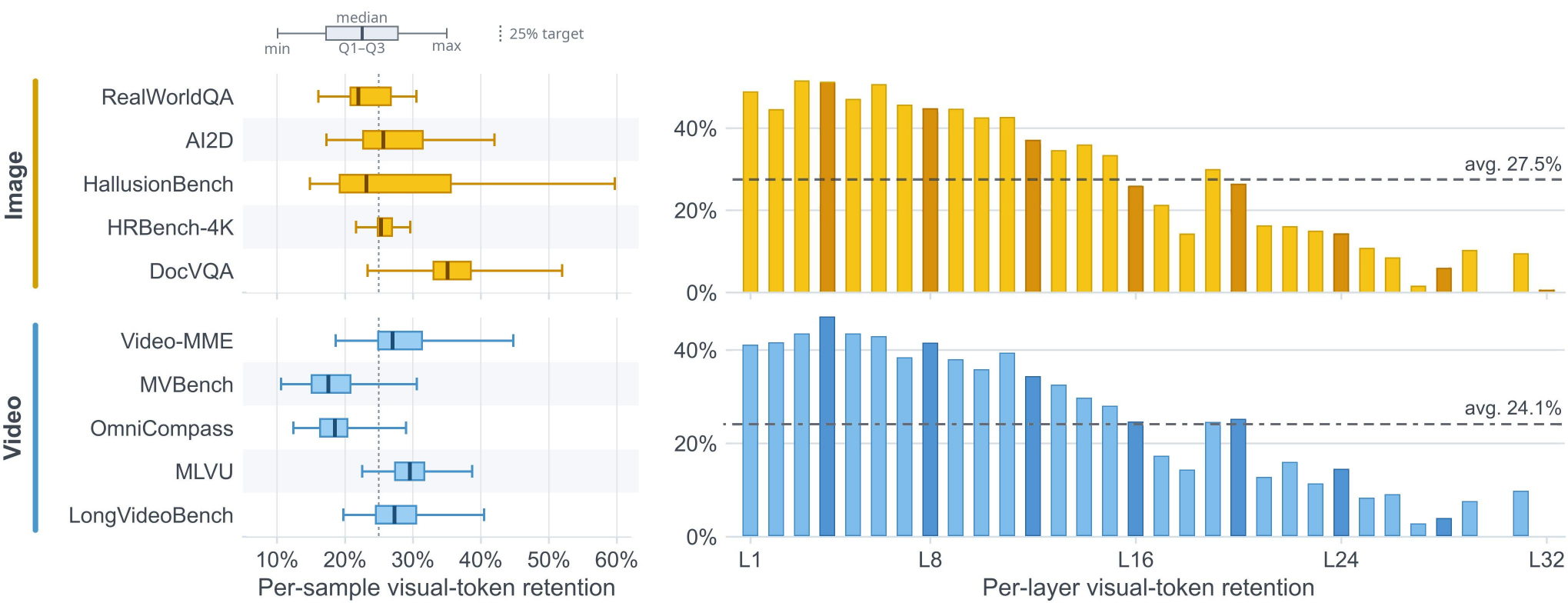}
    \caption{\textbf{SA-MoD adaptively allocates visual tokens across both input samples and model layers on Qwen3.5-4B.}
    \textit{Left:} Per-sample visual-token retention distributions across representative image and video benchmarks under the 25\% target-retention setting (showing min, Q1, median, Q3, and max), revealing substantial content-dependent variation around the target budget.
    \textit{Right:} Layer-wise average visual-token retention over image and video benchmarks across all 32 layers, where dashed lines indicate the global retention. Darker bars at every fourth layer ($\text{L4}, \text{L8}, \dots, \text{L32}$) denote full-attention layers, while others denote linear-attention layers.
    Together, these results confirm that SA-MoD adapts token budgets dynamically across both inputs and network depth rather than enforcing a uniform schedule.}
    \label{fig:amod_retention_dist}
\end{figure}

%% file: figs/vis-cases.tex
\begin{figure}[t]
    \centering
    \includegraphics[width=\linewidth]{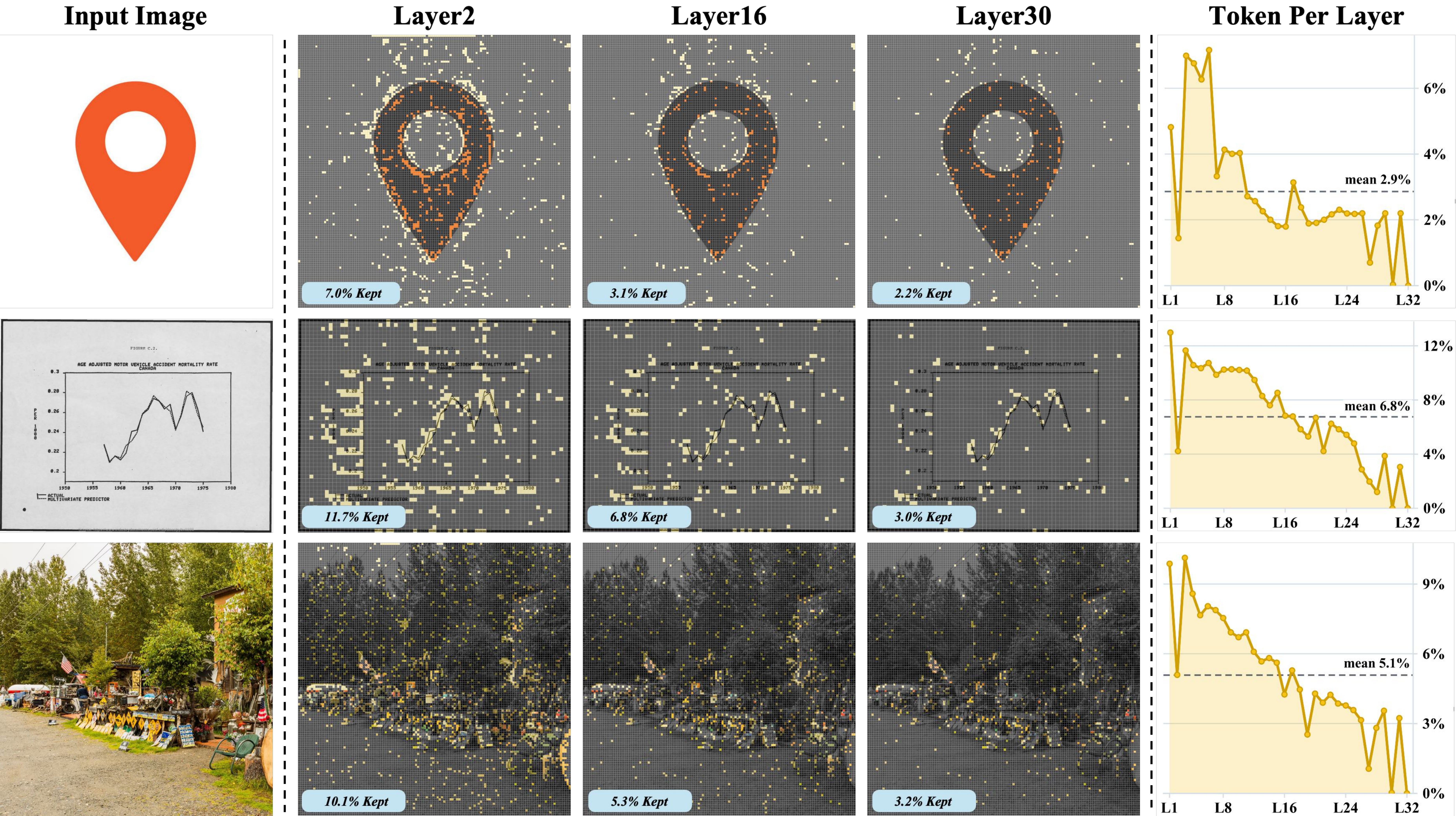}
    \caption{\textbf{SA-MoD exhibits input-adaptive and layer-wise visual-token sparsification.} For three representative inputs of varying complexity, we visualize the retained visual tokens at different Transformer layers alongside their layer-wise retention trajectories. SA-MoD dynamically allocates different token budgets across inputs, while progressively adjusting token retention across network depth. The accompanying curves (\textit{right}) show the non-uniform, input-dependent sparsification trajectories, with dashed lines indicating the average retention across layers. These results highlight SA-MoD's ability to adapt visual computation jointly to input content and network depth.}
    \label{fig:vis_cases}
\end{figure}